\documentclass[runningheads]{llncs}

\usepackage{eccv}

\usepackage{eccvabbrv}

\usepackage{graphicx}
\usepackage{booktabs}
\usepackage{tabu}
\usepackage{xcolor}
\usepackage{color}
\usepackage{colortbl}
\usepackage{subcaption}
\usepackage{multirow}
\usepackage{wrapfig}
\usepackage{tikz}

\usepackage{pifont}
\newcommand{\cmark}{\ding{51}}%
\newcommand{\xmark}{\ding{55}}%
\usepackage{tablefootnote}

\usepackage[accsupp]{axessibility}  %

\usepackage{hyperref}

\usepackage{orcidlink}

\begin{document}

\title{X-Former: Unifying Contrastive and Reconstruction Learning for MLLMs}

\titlerunning{X-Former: Unifying Contrastive and Reconstruction Learning for MLLMs}

\author{Sirnam Swetha\inst{1}\orcidlink{0009-0004-8842-5990} \and
Jinyu Yang\inst{2}\orcidlink{0000-0002-7004-3570} \and Tal Neiman\inst{2}\orcidlink{0009-0005-0198-240X} \and Mamshad Nayeem Rizve\inst{2}\orcidlink{0000-0001-5378-1697} \and \\
Son Tran\inst{2} \and Benjamin Yao\inst{2} \and Trishul Chilimbi\inst{2}\and Mubarak Shah\inst{1,2} }

\authorrunning{S.~Swetha et al.}

\institute{Center for Research in Computer Vision, University of Central Florida\\
\email{swetha.sirnam@ucf.edu, shah@crcv.ucf.edu}
\and Amazon \\
\email{\{viyjy, taneiman, mnrizve, sontran, benjamy, trishulc\}@amazon.com}}

\maketitle

\begin{abstract} 
Recent advancements in Multimodal Large Language Models (MLLMs) have revolutionized the field of vision-language understanding %
by integrating visual perception capabilities into Large Language Models (LLMs).  
The prevailing trend in this field involves the utilization of a vision encoder derived from vision-language contrastive learning (CL),  showing expertise 
in capturing overall representations while facing difficulties in capturing detailed local patterns. 
In this work, we focus on enhancing the visual representations for MLLMs by combining high-frequency and detailed visual
representations, obtained through masked image modeling (MIM), with semantically-enriched low-frequency representations captured by CL. 
To achieve this goal, we introduce X-Former which is a lightweight transformer module designed to exploit the complementary strengths of CL and MIM through an innovative interaction mechanism.
Specifically, X-Former first bootstraps vision-language representation learning and multimodal-to-multimodal generative learning from two frozen vision encoders, i.e., CLIP-ViT (CL-based) and MAE-ViT (MIM-based).
It further bootstraps vision-to-language generative learning from a frozen LLM to ensure visual features from X-Former can be interpreted by the LLM.
To demonstrate the effectiveness of our approach, we assess its performance on tasks demanding detailed %
visual understanding. 
Extensive evaluations indicate that X-Former excels in visual reasoning tasks involving both structural and semantic categories in the GQA dataset.
Assessment on fine-grained visual perception benchmark further confirms its superior capabilities in visual understanding.
\end{abstract}
    
\section{Introduction}
\label{sec:intro}

Recently, Large Language Models (LLMs) have demonstrated remarkable success in diverse natural language tasks \cite{brown2020language,wei2022emergent}, prompting researchers to explore the integration of visual understanding capabilities into these models, leading to multimodal LLMs (MLLMs). 
MLLMs aim to leverage the vast knowledge contained within off-the-shelf LLMs and vision encoders to tackle complex visual understanding tasks, thereby opening up new possibilities in the domain of vision-language understanding. 
Flamingo~\cite{alayrac2022flamingo} is one of the early MLLMs to align frozen visual encoders to LLMs, where it introduces a Perceiver Resampler module to extract a fixed set of features from image by optimizing image-to-text generation loss, in order to bridge the modality gap.
Improving upon Flamingo, BLIP-2~\cite{li2023blip2} proposed a Querying Transformer (Q-Former) that performs vision-language alignment through cross modality fusion by employing both discriminative (contrastive \& classification) and generative (image-to-text generation) losses to extract a fixed set of most useful visual features for LLM.
Other concurrent works~\cite{liu2023llava, zhu2023minigpt} have explored different strategies to align visual representations with LLM input space for improving vision-language understanding.

It is noteworthy that all aforementioned MLLMs employ CLIP-ViT \cite{radford2021clip} as the vision encoder, hence, inherit its limitations including: (i) poor fine-grained vision-language alignment \cite{mukhoti2023openss}, and (ii) spatially-invariant global representations~\cite{park2023whatself}. As a consequence, these models struggle to encode detailed visual nuances, including object orientation, structural intricacies, spatial relationships, and multiple object instances~\cite{tong2024eyes}, thereby hindering the ability of LLMs to comprehend local visual patterns.
To alleviate this issue, there has been growing interest to learn better visual representations for MLLMs. 
For instance, Shikra~\cite{chen2023shikra} proposes to learn visual grounding for objects by adding spatial coordinates in natural language for LLM. However, this requires high-quality curated data with bounding box annotations referring to the objects in the image.

GVT~\cite{wang2023gvt} on the other hand distills features from pre-trained CLIP \cite{radford2021clip} via $L_{1}$ loss and uses the distilled model as the image encoder for extracting visual tokens. However, this approach relies on %
instruction tuning utilizing LLaVA-150k~\cite{liu2023llava} dataset. %
Most recently, MMVP\cite{tong2024eyes} proposes to leverage self-supervised pre-trained vision encoder along with CLIP-ViT to learn Mixture of Features from multiple encoders in LLaVA framework with \textit{LLM fine-tuning}. However, they do instruction tuning with LLaVA-150k~\cite{liu2023llava} dataset. %
Therefore, its not clear whether such an approach can work on commonly available image-text data without relying on instruction tuning using curated datasets.  
An additional avenue of exploration involves constructing a self-supervised vision encoder capable of capturing both global, semantically enriched, and local, detailed visual features. %
The central concept involves linearly combining the training objectives of CL \cite{radford2021clip} and MIM \cite{he2022mae}. This is motivated by the fact that MIM can effectively capture local %
and high-frequency representations, complementing the global %
and low-frequency representations captured by CL. However, this hasn't been explored for vision-laguage understanding and is also the focus of this work.

In this paper, we present X-Former, a lightweight transformer module designed to achieve effective vision-language alignment from both a global and local perspective.
Particularly, X-Former adopts a two-stage training approach. The first stage involves vision-language representation learning and multimodal-to-multimodal generative learning by leveraging two frozen image encoders. Specifically, X-Former utilizes learnable query vectors to extract visual features by utilizing both CLIP-ViT \cite{radford2021clip} and MAE-ViT \cite{he2022mae} encoders as well as employ a dual cross-attention module to dynamically fuse the extracted features. Aimed at image reconstruction and text generation, X-Former is incentivized to extract visual features covering both low frequency %
and high frequency. 

Our main technical contributions can be summarized as: 
\begin{itemize}
\item We propose to leverage vision encoders from CL~\cite{radford2021clip} and MIM~\cite{he2022mae} to capture both global and local visual representations from frozen image encoders to improve vision-language understanding.
\item We introduce X-Former with dual cross-attention to bootstrap multimodal-to-multimodal generative learning using image-text pairs, entirely without the need for curated or visual instruction data.
\end{itemize}

\noindent Empirical studies showcase the notable enhancement of our model in fine-grained visual perception tasks that demand a nuanced understanding of visual details. Specifically, in object counting tasks, X-Former demonstrates substantial improvement over BLIP-2 \cite{li2023blip2} (39.64 vs. 34.3 on COCO and 27.24 vs. 18.9 on VCR). Further, we perform fine-grained analysis comparing the image-text queries of our model and BLIP-2 to demonstrate our approach learns more diverse queries over BLIP-2 indicating the ability to capture detailed %
visual features.
It's worth noting that BLIP-2 is pre-trained on a dataset of 129 Million image-text pairs, approximately $10 \times$ larger than the dataset used for training X-Former (14 Million). This underscores the effectiveness and efficiency of our approach.

\section{Method}
\label{sec:method}
       
In this section, we first briefly recapitulate the preliminaries of Q-Former \cite{li2023blip2}. Following this, we embark on early endeavors aimed at enhancing its visual learning capabilities by leveraging off-the-shelf vision encoders, namely CLIP-ViT and MAE-ViT. Specifically, CLIP-ViT is pre-trained through vision-language contrastive learning strategies, whereas MAE-ViT is trained  %
through masked image modeling mechanisms. Our empirical studies reveal that naively combining these two encoders fails to yield significant performance improvements, especially in tasks necessitating detailed %
visual comprehension. To mitigate this limitation, we introduce a lightweight transformer module, dubbed X-Former, which extends Q-Former to encapsulate both global and local information.

\subsection{Preliminaries of Q-Former}
\begin{figure}[tb]
    \centering
    \begin{subfigure}[]{0.33\textwidth}
        \includegraphics[width=\textwidth]{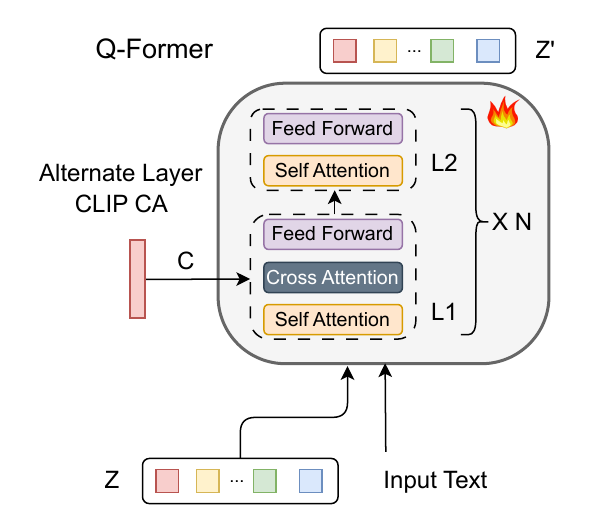}
    \caption{Q-Former}
    \end{subfigure}%
    \begin{subfigure}[]{0.33\textwidth}
        \includegraphics[width=\textwidth]{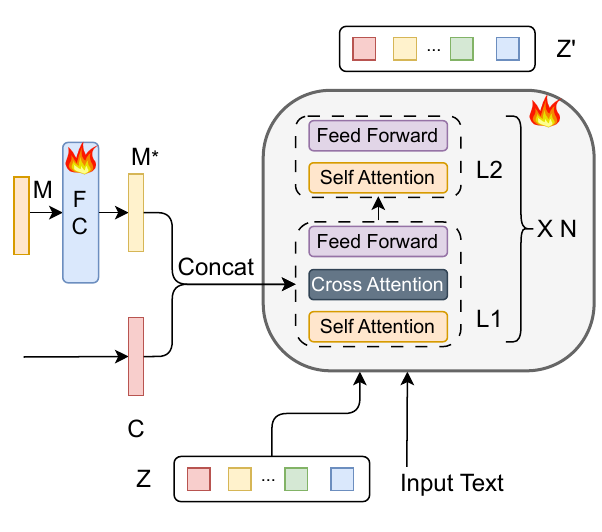}
    \caption{Concatenation }
    \end{subfigure}%
    \begin{subfigure}[]{0.33\textwidth}
        \includegraphics[width=\textwidth]{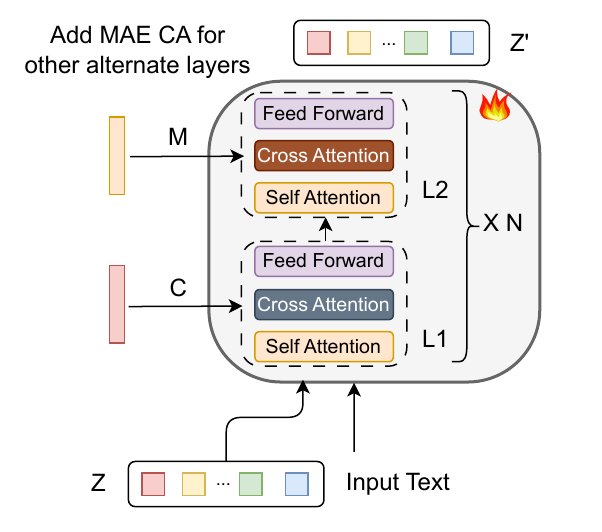}
    \caption{Early Cross-Attn }
    \end{subfigure}%
    \caption{
    \textbf{(a)} Vanilla Q-Former extracts a fixed number of output features $Z'$ from the CLIP image encoder, where $C$ and $z$ denotes CLIP-ViT's image features and the query input, respectively; \textbf{(b)} Concatenated MAE-ViT $(M^{*})$ and CLIP-ViT $(C)$ features are passed as input to Q-Former, \textbf{(c)} A Cross-Attention layer is added in L2 to enable MAE-ViT interaction in Q-Former.}
    \label{fig:overview_s1}
\end{figure}

Q-Former is introduced in BLIP2 \cite{li2023blip2} as a solution designed to bridge the gap between a frozen CLIP-ViT and a frozen LLM (Figure~\ref{fig:overview_s1} (a)). 
Given a collection of image-text pairs $\{(I_{k}, T_{k})\}_{k=1}^{N}$, Q-Former operates by taking a predetermined number of learnable query embeddings $z$, $T_{k}$, and $C$ as input, where $C$ indicates CLIP image features of $I_{k}$.
These queries engage in mutual interaction through self-attention layers and interact with frozen image features $C$ through cross-attention layers in every alternate  layer as shown in Figure~\ref{fig:overview_s1}(a) L1. 
The resulting query representation is denoted by $Z'$, which is anticipated to encapsulate visual information derived from the frozen CLIP-ViT.

Though Q-Former has exhibited remarkable performance on various downstream tasks like VQA and image captioning, it encounters challenges in detailed %
visual feature comprehension. This limitation primarily stems from the training objective of CLIP, which incentivizes ViT to prioritize low-frequency signals and global visual patterns \cite{park2023whatself}. Fortunately, MAE-ViT \cite{he2022mae}, trained to reconstruct masked image patches, excels in understanding detailed %
visual features. However, the integration of CLIP-ViT and MAE-ViT in multimodal understanding remains unclear, given their inherently divergent perspectives when `viewing' images. To address this inquiry, we embark on early attempts to combine CLIP-ViT and MAE-ViT in a straightforward manner as discussed below. %

\subsection{Simple Combinations of CLIP-ViT and MAE-ViT}
\begin{figure}[t]%
    \centering
    \begin{subfigure}[]{0.3\textwidth}
        \includegraphics[width=\textwidth]{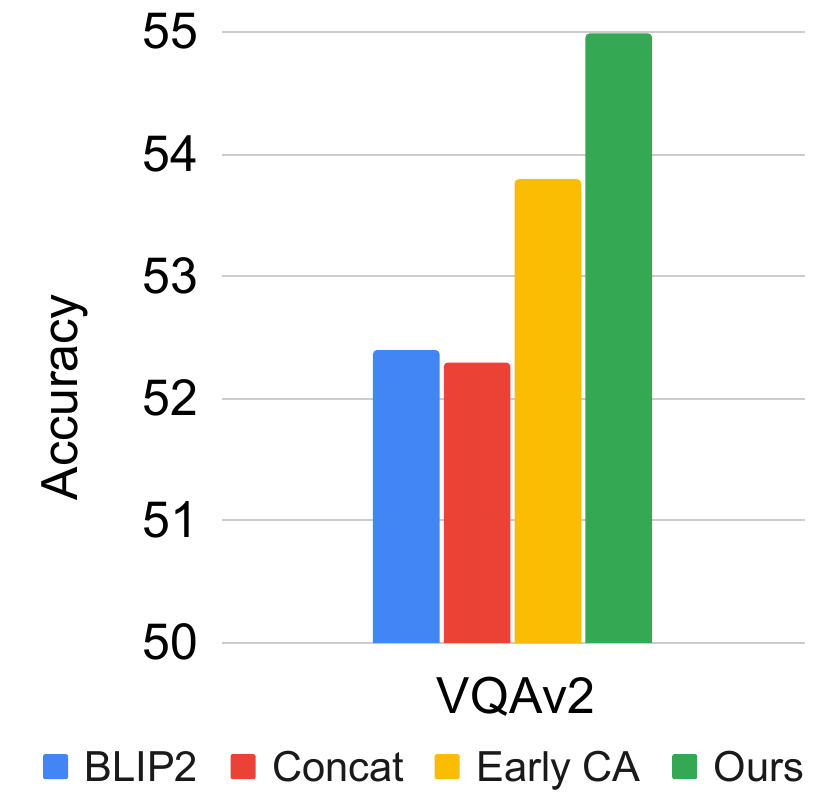}
    \caption{}
    \end{subfigure}%
    \begin{subfigure}[]{0.3\textwidth}
        \includegraphics[width=\textwidth]{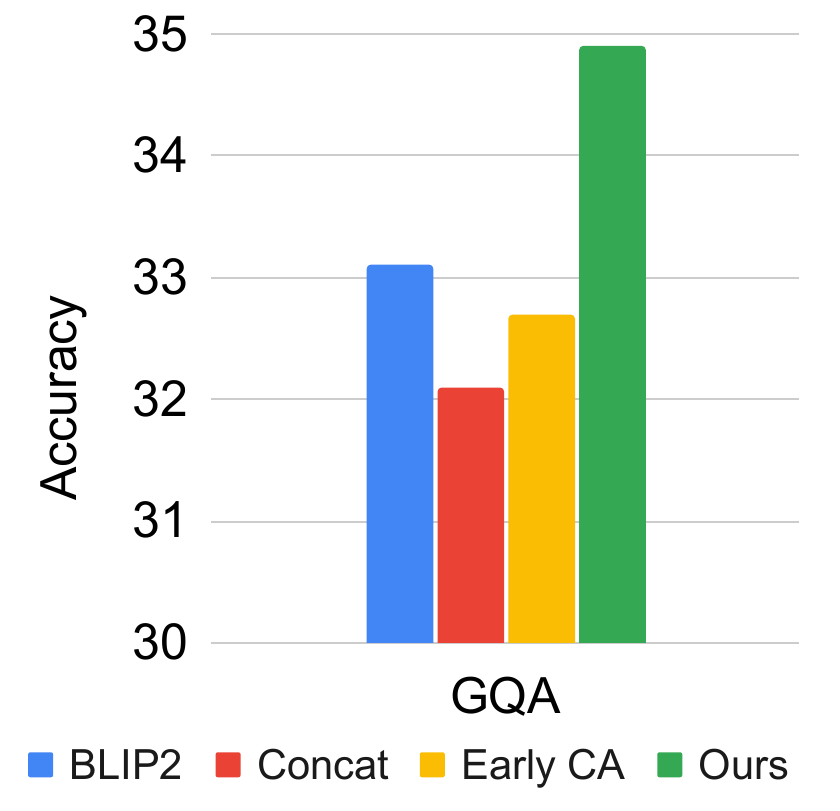}
        \caption{}
    \end{subfigure}%
    \begin{subfigure}[]{0.3\textwidth}
        \includegraphics[width=\textwidth]{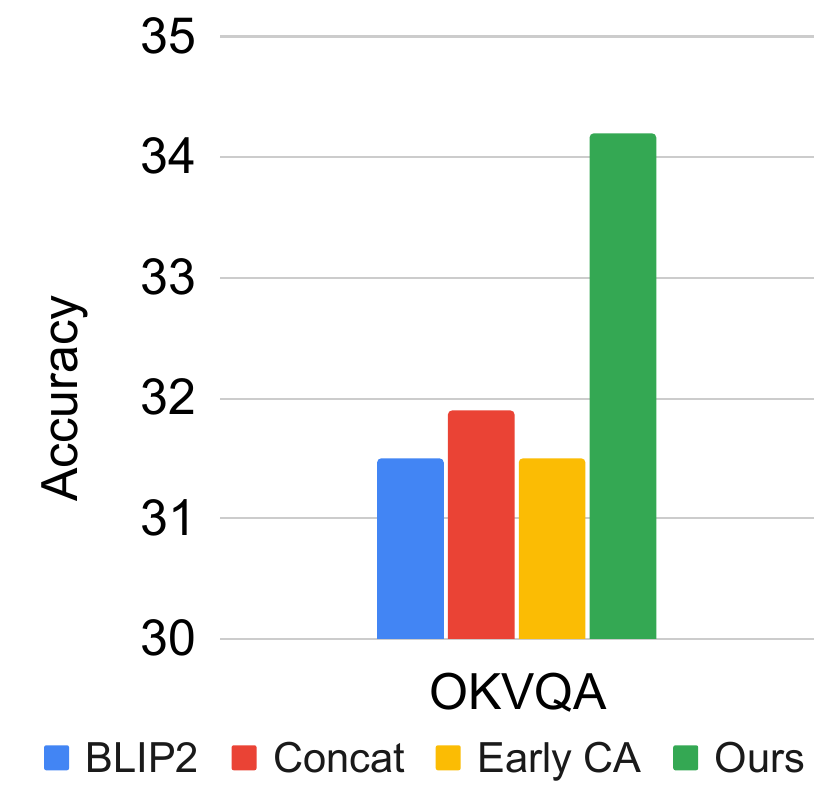}
        \caption{}
    \end{subfigure}%
  \caption{Performance comparison of BLIP2, BLIP2+Concatenation, BLIP2+Early Cross-Attention, and our method on VQAv2 (a), GQA (b), and OKVQA (c) datasets.}
  \label{fig:model_strat_comp}
\end{figure}

\paragraph{\textbf{Visual Feature Concatenation}}
As shown in Figure~\ref{fig:overview_s1} (b), our first attempt is to concatenate the frozen image features from CLIP-ViT and MAE-ViT, which are denoted by $C$ and $M$, respectively. 
To accommodate the discrepancy between $C$ and $M$, a linear layer is applied to align $M$ with $C$, resulting in $M^*$, which is subsequently concatenated with $C$. This combined feature $(C, M^*)$ serves as input to the Q-Former, which undergoes training in both stages following the methodology outlined in~\cite{li2023blip2}. Our experiments show that the simple concatenation approach performs on par with BLIP-2, as illustrated in Figure~\ref{fig:model_strat_comp}. This observation highlights the non-trivial nature of integrating $C$ and $M$ to leverage their complementary strengths. The distinct information provided by MAE and CLIP presents challenges for the model in simultaneously learning both global and local information while preserving visual-text coherence. Moreover, it is crucial to note that introducing additional vision encoders does not necessarily guarantee improved performance.

\paragraph{\textbf{Early Cross-Attention}}
Inspired by the observations from the concatenation strategy outlined earlier, we delve into early interactions akin to CLIP-style cross-attention within Q-Former. To pursue this, we introduce early cross-attention by integrating new cross-attention layers, alternating with non-CLIP interaction layers, as depicted in Figure~\ref{fig:overview_s1} (c). While this approach modestly improves performance compared to the concatenation strategy (see Figure~\ref{fig:model_strat_comp}), it notably escalates the number of parameters in Q-Former, resulting in a total of 183M trainable parameters (approximately 75M more than BLIP-2). Importantly, increasing parameters doesn't inherently enhance performance. While enhancements are observed for the VQAv2 dataset, there's a decline in performance for the GQA dataset and comparable results for the OKVQA dataset against BLIP-2. To mitigate this and facilitate the extraction of local information from MAE, we advocate for incorporating late-interaction for the Masked Image Modeling (MIM) objective during training.
\subsection{X-Former Overview}
\label{sec:method_details}
\begin{figure*}[tb]
    \centering
    \includegraphics[width=0.94\textwidth]{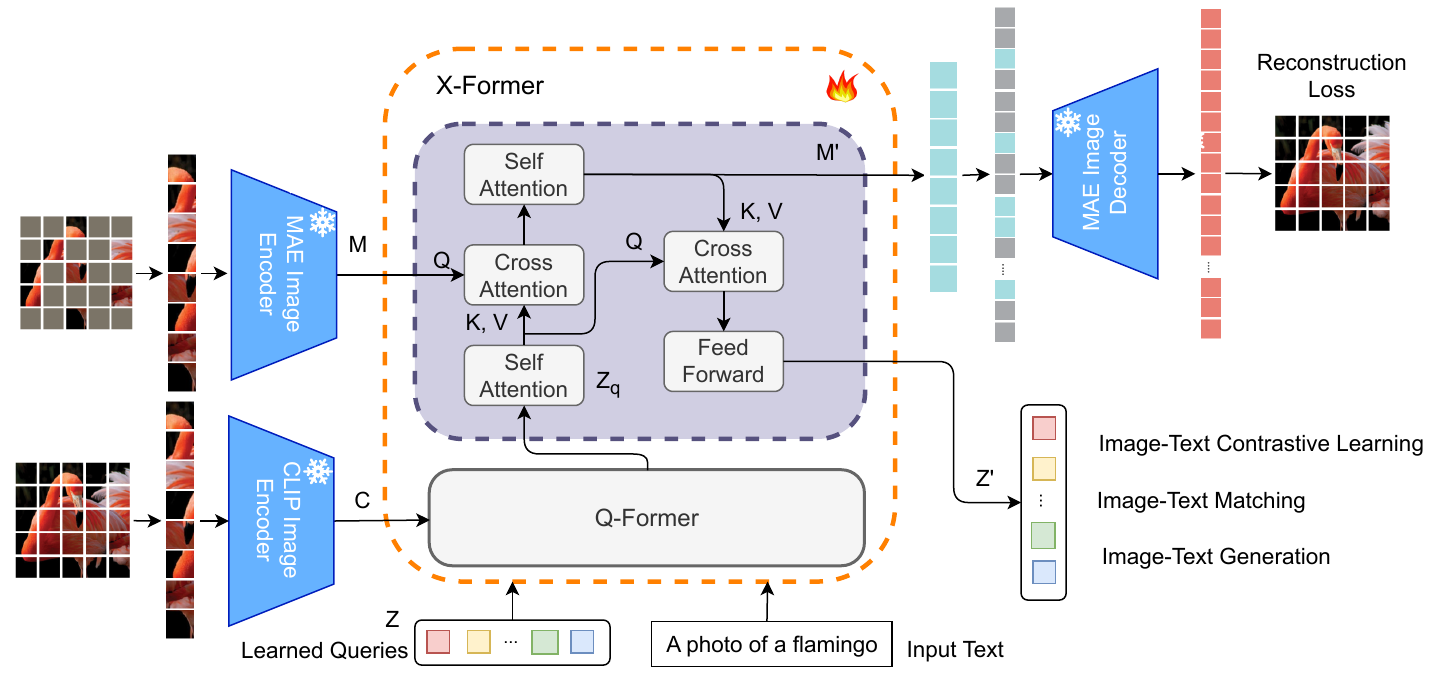}
    \caption{
    An overview of X-Former which extends Q-Former by introducing a dual cross-attention module to capture both local and global visual features. %
    First it computes CLIP visual features $(C)$ and MAE features $(M)$ (with random masking) from the input image-text pair.
    Q-Former employs $C, Z,$ Text to generate output queries optimized for three objectives - ITC, ITM and ITG. The proposed block (purple) enriches Q-Former global representation $(Z_q)$ with local information from MAE features $(M)$. Initially, $M$ is aligned and enriched by $Z_q$ resulting in enriched MAE representation $(M')$, optimized for image reconstruction. Then, $M'$ enhances $Z_q$ with local representations through cross-attentions, optimized using VL objectives. Jointly optimizing these four objectives facilitates the learning of both global and local representations.}
    \label{fig:sys_s1}
\end{figure*}

In Figure~\ref{fig:sys_s1}, we present an overview of our method, comprising two frozen image encoders (CLIP-ViT and MAE-ViT), a frozen image decoder, and a trainable X-Former aimed at bridging the modality gap and extracting interpretable visual features for the LLM. For MAE-ViT, random masking of patches in the input image is performed. X-Former processes a set of learnable queries $Z$ along with the input text $T_{k}$ and the image features $(C, M)$ as input. Our model extends the framework of BLIP2 by incorporating Image-Text Matching (ITM), Image-Text Contrastive (ITC), and Image-Text Generation (ITG) losses, while also introducing a reconstruction loss for the image decoder.
\paragraph{\textbf{X-Former}} 
To address the limitations of Q-Former, primarily its lack of fine-grained alignment and its focus on capturing global information, we propose integrating MAE features $(M)$ into our X-Former module, depicted as an orange block in Figure~\ref{fig:sys_s1}. This addition facilitates the extraction of both local and global information, optimizing image reconstruction alongside the ITC, ITM, and ITG objectives, represented by the purple block in Figure~\ref{fig:sys_s1}. The first cross-attention block employs MAE features $(M)$ as queries and Q-Former output $(Z_q)$ as keys and values to align and enhance $M$ by integrating global semantic information from Q-Former, resulting in enriched MAE features $(M')$. Subsequently, these enriched MAE features enhance the Q-Former output $(Z_q)$ to $Z'$ by integrating both global and local information through cross-attention, as depicted. The enhanced queries $(Z')$ are optimized for ITC, ITM, and ITG, along with a reconstruction objective applied to $M'$. Finally, $M'$ is passed to the frozen MAE decoder to reconstruct the masked patches. 

\paragraph{\textbf{Stage 1: Pre-Training}}
\label{sec:s1_pre}
During the pre-training stage, the X-Former learns to extract both local and global representation by optimizing Reconstruction, ITC, ITM and ITG losses. The reconstruction loss together with the image-text alignment objectives enforces to align and capture local representation, while the VL objectives align it with text representation. The incorporation of MAE and CLIP features ensures that the queries extract a enhanced visual representation that is aligned with the accompanying text. We follow BLIP-2~\cite{li2023blip2} for computing ITC, ITM and ITG losses. 
For ITC, we compute similarity between [CLS] token of the text-embedding and each of the final output query embeddings $Z'$, selecting the \textit{highest} as the image-text similarity. For this objective, to prevent data leak a unimodal self-attention mask is employed, ensuring that the queries and text do not interact with each other. It maximizes the image-text similarity of positive pairs by contrasting with in-batch negatives. 
  
For ITM, the model is asked to predict whether image-text pair match (positive) or not (negative). Here, a bi-directional self-attention mask is employed, allowing all queries and texts to attend to each other. Consequently, the output query embeddings capture multimodal information, which is then fed to a two-class linear classifier to obtain logits. These logits are averaged across all the queries to compute the final matching score. To generate negative pairs, a hard negative mining strategy~\cite{li2022blip} is employed. 
In the context of ITG, X-Former utilizes an input image as a condition to generate text. A multimodal causal self-attention mask is used, allowing queries to attend to each other while excluding text tokens, and enabling text tokens to attend to all queries and previous text tokens. The [CLS] token is substituted with the [DEC] token as the first text token, serving as an indicator for the decoding task.
\begin{figure}[tb]
  \centering
    \includegraphics[width=0.75\textwidth, clip=true]{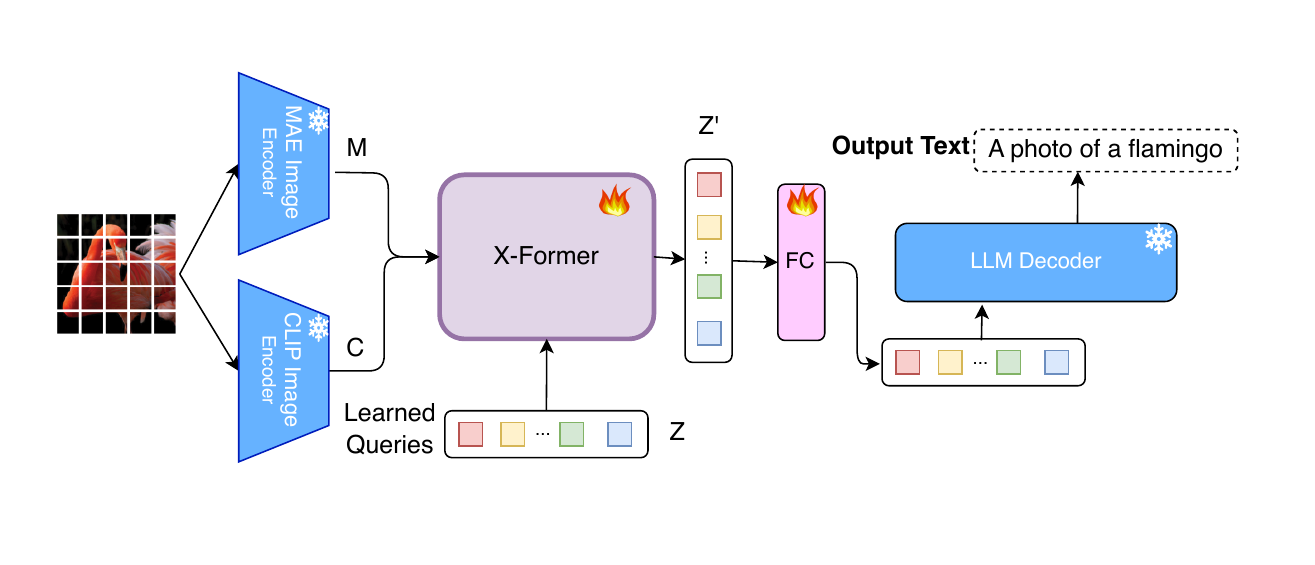}
  \caption{LLM Alignment. X-Former queries are aligned with a \textit{frozen} decoder-based LLM. FC layer adapts the query output$(Z^{\prime})$ to LLM embedding space.
  }
  \label{fig:stage2_overview}
\end{figure}

\paragraph{\textbf{Stage 2: LLM Alignment}}
\label{sec:s2_llm}
During pre-training, the X-Former acquires the  
ability to extract information from both MAE and CLIP, resulting in queries that capture a blend of global and local information. Subsequently, we align the features of the X-Former with the frozen LLM, aiming to harness the comprehensive visual representations acquired by the X-Former module and integrate them with the robust language generation capabilities of the LLM. This integration involves connecting the pre-trained X-Former output ($Z'$) to the LLM via a single fully-connected layer, aligning it with the LLM representation space, as depicted in Figure~\ref{fig:stage2_overview}. Specifically, we experiment with the OPT model, which is a decoder-based LLM, and train it using a language modeling loss keeping \textit{both} image encoders and LLM \textit{frozen}.

\section{Experiments}
\label{sec:exp}
\paragraph{\textbf{Pre-Trained Models}} 
We employ pre-trained ViT-G model from EVA-CLIP~\cite{fang2023eva} as CLIP-ViT. 
For MAE, we utilize the pre-trained ViT-H model ~\cite{he2022mae}.
Our choice for the LLM involves the OPT model~\cite{zhang2022opt}. Our model undergoes pre-training for nine epochs in Stage-1 and one epoch in Stage-2, with OPT employed for Stage-2 alignment. %
See Supplementary Section 1 for implementation details.
\begin{table*}[tb]\setlength{\tabcolsep}{10pt}
  \centering
  \caption{Zero-shot Visual Question Answering results on the VQAv2 dataset. Note that * indicates the result is obtained using the official checkpoint.}
  \resizebox{0.96\textwidth}{!}{
  \begin{tabu}{lcccccc}
    \toprule
    & $\#$Trainable & & \multicolumn{4}{c}{VQAv2 Accuracy}                   \\
       Method & Params & Data & Overall & Other & Yes$/$No & Number  \\
    \midrule
    & & & \multicolumn{4}{l}{Open-ended generation models}\\
    \midrule
    FewVLM~\cite{fewvlmjin-etal-2022-good} & 740M & 9.1M & 47.7 & - & - & -   \\
    Frozen~\cite{Frozen_NEURIPS2021_01b7575c} & 40M & & 29.5 & - & - & -   \\
    VLKD~\cite{VLKD_dai-etal-2022-enabling} & 406M & 3.7M & 42.6 & - & - & -   \\
    \rowfont{\color{gray}}
    BLIP-2 $OPT_{6.7B}$*~\cite{li2023blip2} & 108M & 129M & 55.1 & 47.3 & 72.6 & 34.6  \\
    \hline
    BLIP-2 $OPT_{2.7B}$~\cite{li2023blip2} & 107M & 14M & 49.9 & 39.3 & 71.5 & 27.3  \\
    \textbf{X-Former (Ours)} $\mathbf{OPT_{2.7B}}$ & 129M & 14M & \textbf{51.3} & \textbf{41.5} & 71.2 & \textbf{30.9}  \\
    BLIP-2 $OPT_{6.7B}$~\cite{li2023blip2} & 108M & 14M & 52.4 & 43.6 & 71.5 & 30.8  \\
    \textbf{X-Former (Ours)} $\mathbf{OPT_{6.7B}}$ & 130M & 14M & \textbf{55.0} & \textbf{45.6} & \textbf{73.3} & \textbf{37.8}  \\
    \bottomrule
  \end{tabu}
  }
  \label{tab:zr_s2_vqa}
\end{table*}
\paragraph{\textbf{Datasets and Tasks}}
To demonstrate the effectiveness of our approach, we leverage a standard dataset comprising 14M Image-Text pairs sourced from COCO~\cite{lin2014microsoftcoco}, Visual Genome~\cite{krishna2017visualgenome}, SBU~\cite{ordonez2011im2textsbu}, CC3M~\cite{sharma2018conceptual3m}, and CC12M~\cite{changpinyo2021conceptual12m} for model pre-training. %
Our evaluation spans across various benchmarks, including COCO~\cite{lin2014microsoftcoco}, NoCaps~\cite{agrawal2019nocaps}, VQAv2~\cite{goyal2017making}, GQA~\cite{hudson2019gqa}, OK-VQA~\cite{marino2019ok}, Flickr30k~\cite{plummer2015flickr30k}, and VCR~\cite{zellers2019vcr}.
Furthermore, we employ a fine-grained visual perception benchmark~\cite{wang2023gvt}, featuring Object Counting (OC) and Multi-Class Identification (MCI) tasks, to assess the model's fine-grained visual understanding capabilities. %

\subsection{Experimental Results}
\paragraph{\textbf{Zero-Shot Visual Question Answering}}
First, we present the results for zero-shot visual question answering on the VQAv2-val dataset, which encompasses three question types: open-ended (other), Yes/No, and Number questions, as illustrated in Table~\ref{tab:zr_s2_vqa}. We utilize the prompt "Question: ${}$ Short Answer:" for the generation process, employing beam search with a beam width of 5. We set the length-penalty to 0 to encourage short answers. Our results indicate that our approach surpasses BLIP-2 for both $OPT_{2.7B}$ and $OPT_{6.7B}$ LLMs by 1.4\% and 2.6\% respectively, highlighting superior visual comprehension. Particularly noteworthy are the significant enhancements observed for the Number task, which demands precise local understanding for object counting or identification. Fine-tuning results are reported in Supplementary Section 2, while large-scale experimental findings are detailed in Supplementary Section 3.
\begin{table}[t]\setlength{\tabcolsep}{5pt}
  \centering
  \caption{Zero-shot Visual Question Answering Results on GQA and OKVQA datasets. Note that * indicates the result is obtained using the official checkpoint.}
  \begin{tabu}{lccc}
    \toprule
       Method & Data & GQA & OKVQA  \\
    \midrule
    FewVLM~\cite{fewvlmjin-etal-2022-good} & 9.1M & 29.3 & 16.5 \\
    Frozen~\cite{Frozen_NEURIPS2021_01b7575c} &  & - & 5.9 \\
    VLKD~\cite{VLKD_dai-etal-2022-enabling} & 3.7M & - & 13.3 \\
    Flamingo3B~\cite{alayrac2022flamingo} & $>$2B &-  & 41.2 \\
    Flamingo9B~\cite{alayrac2022flamingo} & $>$2B & - & 44.7 \\
    Flamingo80B~\cite{alayrac2022flamingo} & $>$2B & - & 50.6 \\
    \rowfont{\color{gray}}
    BLIP-2 $OPT_{6.7B}$*~\cite{li2023blip2}& 129M & 34.2 & 35.3  \\
    \hline
    BLIP-2 $OPT_{2.7B}$~\cite{li2023blip2} & 14M & 33.6 & 24.2  \\
    \textbf{X-Former (Ours)} $\mathbf{OPT_{2.7B}}$ & 14M & \textbf{34.1} & \textbf{27.7} \\
    BLIP-2 $OPT_{6.7B}$~\cite{li2023blip2} & 14M & 33.1 & 31.5  \\
    \textbf{X-Former (Ours)} $\mathbf{OPT_{6.7B}}$& 14M & \textbf{34.9} & \textbf{34.2} \\
    \bottomrule
  \end{tabu}
  \label{tab:zr_s2_gqa}
\end{table}
\par
\begin{figure*}[tb]
  \centering
    \begin{subfigure}[]{0.49\textwidth}
        \includegraphics[width=\textwidth]{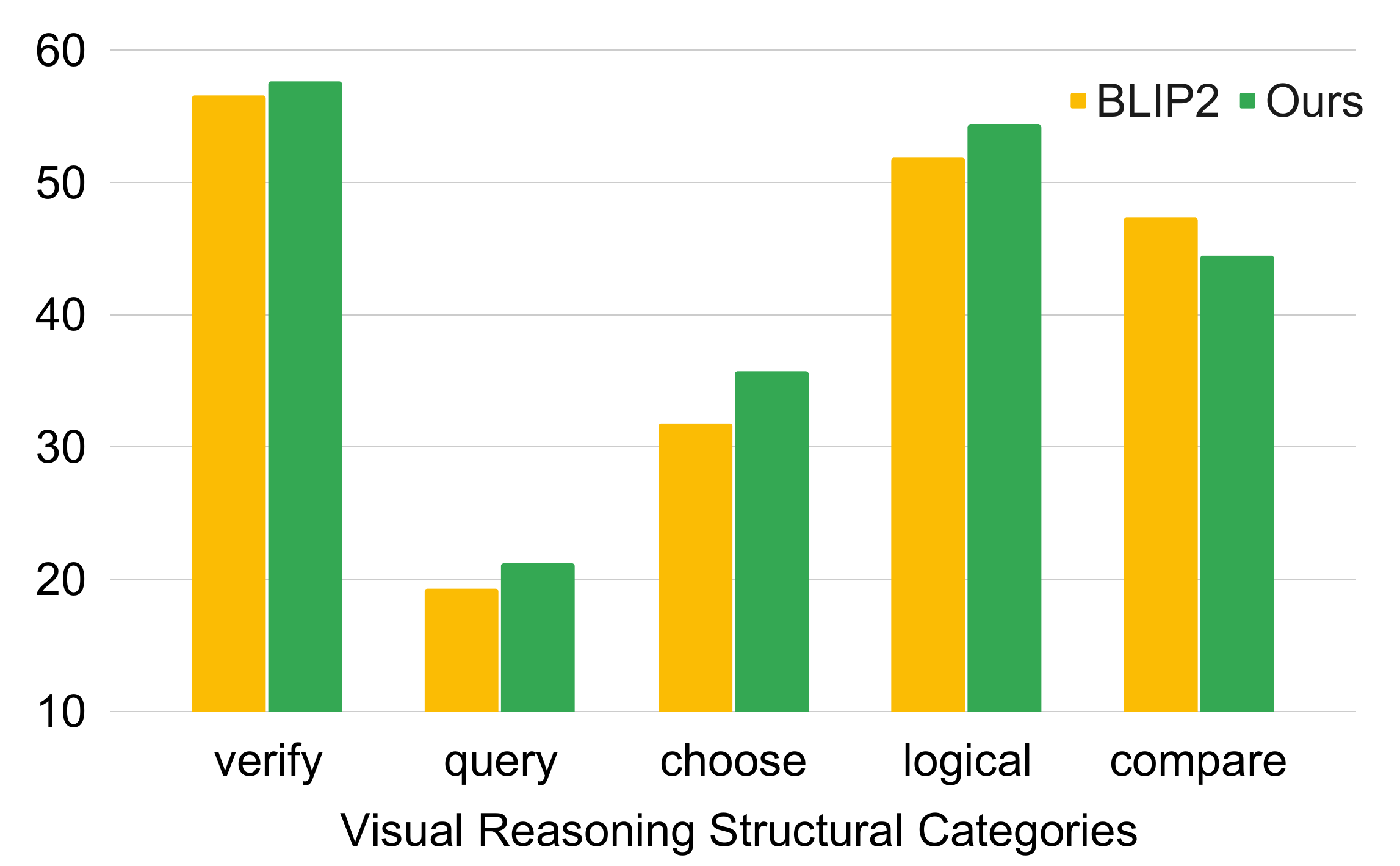}
    \caption{Comparison for Structural categories}
    \end{subfigure}%
    \begin{subfigure}[]{0.49\textwidth}
        \includegraphics[width=\textwidth]{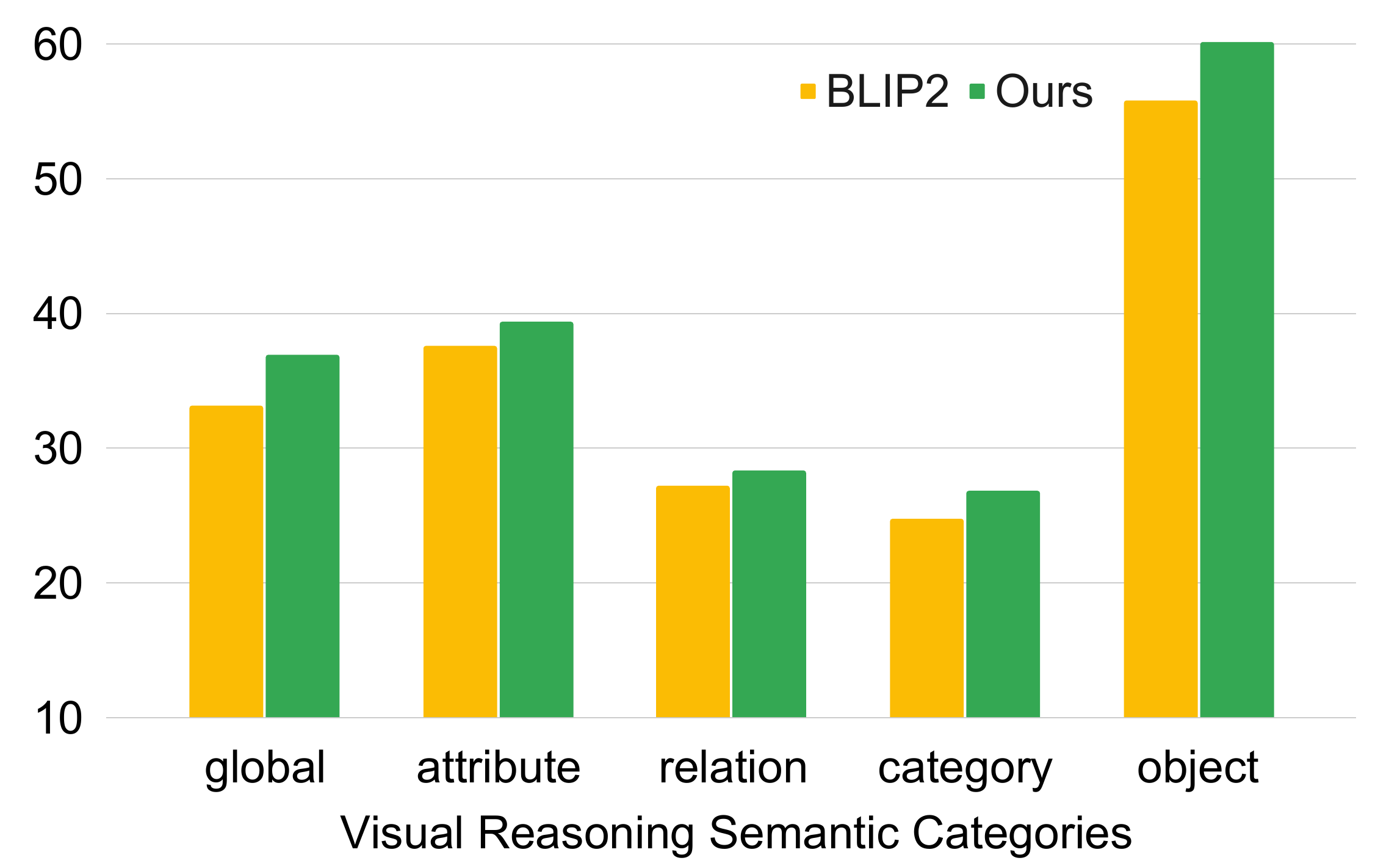}
        \caption{Comparison for Semantic categories}
    \end{subfigure}%
      \caption{Detailed Comparison for both Structural and Semantic categories in GQA.} 
  \label{fig:gqa_ss}
\end{figure*}

 In Table~\ref{tab:zr_s2_gqa}, we report zero-shot visual question answering results for the GQA test-dev dataset. The results  demonstrate the superior performance of our method over BLIP-2. Furthermore, we conducted a comprehensive comparison to demonstrate the effectiveness of our approach across both structural and semantic categories in the GQA dataset. The structural category encompasses five question types (verify, open-ended query questions, choose from options, logical inference, and object comparison) as depicted in Figure~\ref{fig:gqa_ss}. Our results indicate that we outperform in the majority of these categories. In Figure~\ref{fig:gqa_ss}, we provide a comparison for the semantic categories, which include questions related to object existence, object attributes, object category, global scene, and object relationships. Across all these categories which includes both global and local reasoning, our approach consistently demonstrates better performance, highlighting its detailed %
visual understanding capabilities.

\par We report zero-shot visual question answering performance on OKVQA test dataset in Table~\ref{tab:zr_s2_gqa}. This dataset poses a significant challenge as it requires methods to draw upon external knowledge to answer questions effectively. Our method demonstrates a significant improvement in accuracy over BLIP-2, achieving a 2.7\% and 3.5\% gain with $OPT_{6.7B}$ and $OPT_{2.7B}$ LLM respectively.This signifies the robustness of our approach in accurately aligning visual information with LLM and effectively leveraging external knowledge to answer the questions.

\paragraph{\textbf{Fine-Grained Visual Perception Evaluation}} 
To demonstrate that our approach has better visual understanding, we evaluate perception abilities at fine-grained scale~\cite{wang2023gvt}, we evaluate our approach for fine-grained visual perception capabilities OC and MCI task. %
We use the prompt ``Question: $\{\}$ Short Answer:''. to evaluate for this task. For generation, we use beam search with a beam width of 5. We also set the length-penalty to 0 to encourage shorter answers. The questions for object counting tasks is of the form ``How many $\{objects\}$ are there in the image?'' and for multi-class identification task it is ``Does $\{objects\}$ exist in the image?''. For fair comparison, we compare with methods that only employ image-text datasets for training. In Table~\ref{tab:zr_s2_fg}, we show that our model outperforms BLIP-2 on both datasets i.e., COCO and VCR. It can be seen that for Object Counting task our approach improves BLIP-2 by 13\% on COCO and 6.1\% on VCR datasets respectively. 
This indicates that X-Former is able to extract detailed %
visual features. %
Please refer to Supplementary Sections 6, 7 for more fine-grained evaluations. %

\begin{table}[tb]\setlength{\tabcolsep}{10pt}
\caption{Zero-shot Fine-Grained Visual Perception evaluation of MLLMs on Object Counting (OC) \& Multi-class Identification (MCI) tasks. For fair comparison, we compare with models trained only on image-text data. *evaluated using official checkpoint.}
  \centering
  \begin{tabu}{lccccc}
    \toprule
    &  & \multicolumn{2}{c}{OC} & \multicolumn{2}{c}{MCI}                   \\
    \cmidrule(r){3-6}
       Method & Data & COCO & VCR & COCO & VCR  \\
    \midrule
    \rowfont{\color{gray}}
    BLIP-2*~\cite{li2023blip2} & 129M & 34.3 & 18.9 & 69.44 & 74.16  \\
    \hline
    BLIP-2~\cite{li2023blip2} & 14M & 25.88 & 21.12 & 61.5 & 65.3 \\
    \textbf{X-Former (Ours)} & 14M & \textbf{39.64} & \textbf{27.24} & \textbf{69.44} & \textbf{69.28} \\
    \bottomrule
  \end{tabu}
  \label{tab:zr_s2_fg}
\end{table}

\paragraph{\textbf{Zero-shot Image Captioning.}} In addition to the visual reasoning tasks, we report results for image captioning without fine-tuning in Table~\ref{tab:zr_s2_cap} for COCO and NoCaps dataset. Captioning task requires image-level semantic understanding as the annotated captions briefly describe the image. We show that our approach improves on fine-grained visual reasoning tasks without impacting the captioning performance.

\begin{table}[tb]\setlength{\tabcolsep}{7pt}
    \caption{Zero-shot Image Captioning Results on COCO \& NoCaps without fine-tuning for captioning task. B:BLEU, C: CIDEr, S: SPICE. *evaluated using official checkpoint}
  \centering
  \resizebox{0.8\textwidth}{!}{
  \begin{tabu}{lcccccc}
    \toprule
    & & \multicolumn{3}{c}{COCO}  &   \multicolumn{2}{c}{NoCaps}  \\ %
    \cmidrule(r){3-7}
    Method  & Data & B@4 & C & S & C & S \\ %
    \midrule
    \rowfont{\color{gray}}
    BLIP-2*~\cite{li2023blip2} & 129M & 39.9 & 134.3 & 24.3 & 113.4  & 15.2\\ %
    \hline
    BLIP-2~\cite{li2023blip2} & 14M & 39.2 & 131.0 & 23.7 & 113.1  & 14.9 \\ %
    \textbf{X-Former (Ours)} & 14M & \textbf{39.3} & \textbf{131.1} & 23.6 & \textbf{113.2} & 14.9 \\ %
    \bottomrule
  \end{tabu}
  }
  \label{tab:zr_s2_cap}
\end{table}

\begin{figure*}[tb]
    \centering
    \includegraphics[width=0.95\textwidth]{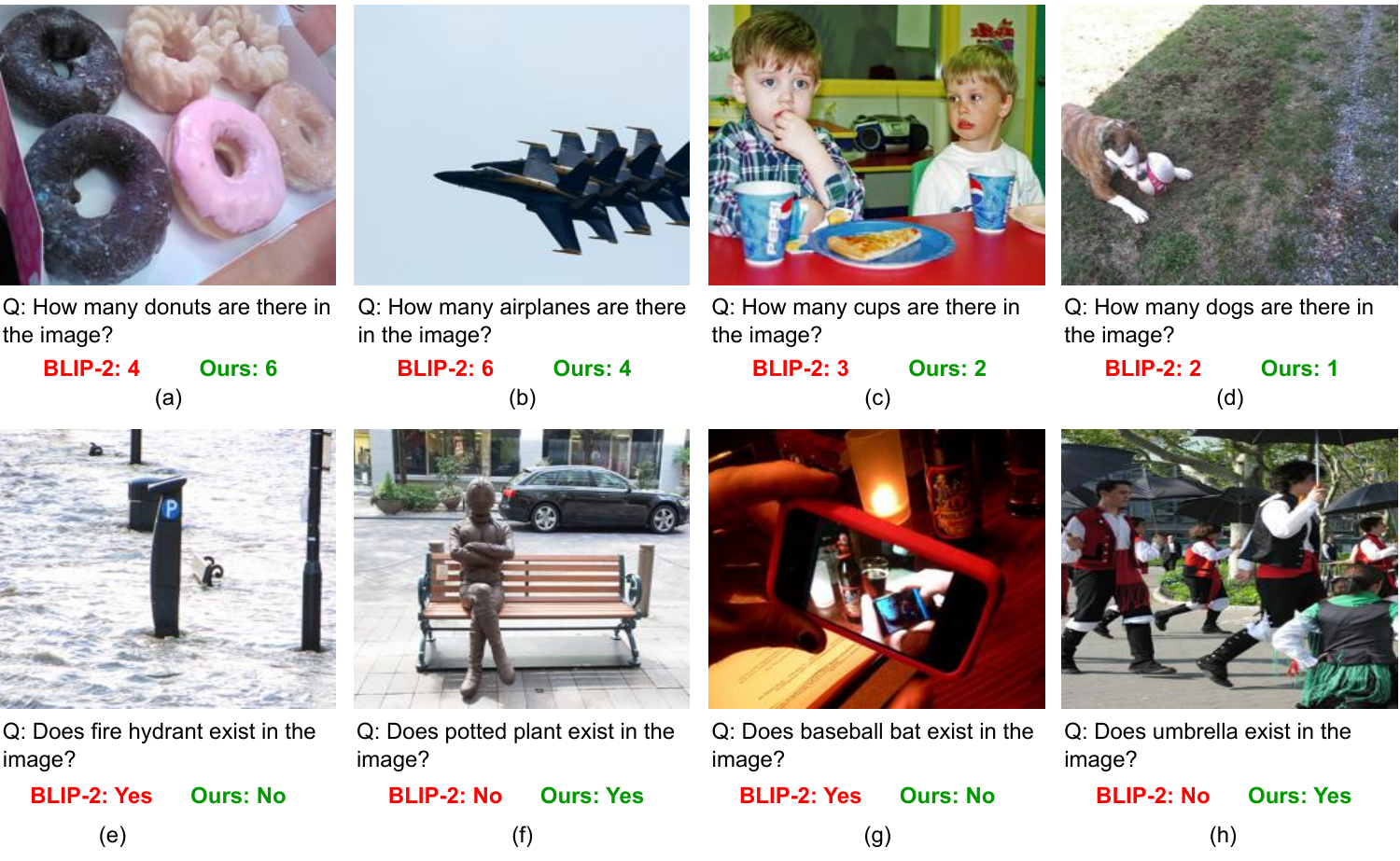}
      \caption{Qualitative Comparison demonstrating Fine-Grained Visual Understanding in Object Counting and Multi-class Identification Tasks. Our model showcases better visual understanding by accurately counting objects and effectively identifying them without confusion based on shape or color.}
  \label{fig:qual}
\end{figure*}

\subsection{Qualitative Results}
To effectively demonstrate the capabilities of our model, we present qualitative results that highlight its performance in the object counting task. Accurate object counting requires a deep understanding of local contexts within an image. As shown in Figure~\ref{fig:qual}(a), our method correctly counts six donuts in an image, while BLIP-2 incorrectly predicts four. This demonstrates the model's ability to distinguish individual objects even when they are closely clustered together. Figure~\ref{fig:qual}(b) presents a more challenging scenario where four airplanes are flying in close proximity. BLIP-2 struggles with this task, erroneously predicting six airplanes instead of the correct number of four. Our method, on the other hand, accurately identifies the four airplanes, showcasing its robustness in handling dense object arrangements.
In Figure~\ref{fig:qual}(c), we encounter two cups and a plate that share a similar color, for which BLIP-2 incorrectly predicts three cups. Our method, however, correctly identifies the two cups, demonstrating its ability to handle objects with similar visual properties.
Figure~\ref{fig:qual}(d) depicts a dog that blends into the background due to its similar coloring. BLIP-2 makes an incorrect prediction. Our method, in contrast, correctly identifies the dog.
These qualitative results collectively demonstrate the effectiveness of our model in the object counting task, outperforming BLIP-2 in various scenarios that demand robust local understanding and the ability to handle challenging object arrangements and color similarities.

In the Multi-Class Identification task, BLIP-2's object recognition capabilities exhibit limitations when presented with Figure~\ref{fig:qual}(e). BLIP-2 mistakenly interprets the shape of a parking pole as a fire hydrant. Figure~\ref{fig:qual}(f) presents a challenge where a potted plant is positioned in the background, occupying a relatively small portion of the image. BLIP-2 fails to detect the presence of the potted plant, whereas our method successfully identifies it. Figure~\ref{fig:qual}(g) showcases image of a bottle whose shape closely resembles that of a baseball bat, leading BLIP-2 to identify it as a baseball bat. Figure~\ref{fig:qual}(h) presents a challenge due to the subtle color of an umbrella, making it difficult task. BLIP-2 fails to recognize the object, while our method accurately identifies it as an umbrella. A comprehensive qualitative analysis is provided in Supplementary Section 5. We also present query diversity analysis for fine-grained qualitative comparison and present them in Supplementary Section 6.

\begin{table}[t]
\centering
\begin{minipage}{.39\linewidth}
\caption{Ablation study with early layer features from CLIP-ViT. $L_{i}$ indicates $i^{th}$ layer of CLIP-ViT.}
\begin{tabular}{lccc}
    \toprule
       Method & VQAv2 & GQA & OKVQA  \\
    \midrule
    Ours & \textbf{55.0} &  \textbf{34.9} &\textbf{34.2}  \\
    $L_{26}$ & 53.7 & 32.6 & 31.2  \\
    $L_{28}$ & 52.5 & 31.9 & 28.0 \\
    $L_{30}$ & 52.4 & 32.8 &  30.9 \\
    \bottomrule
  \end{tabular}
  \label{tab:abl_interim}
\end{minipage}
\begin{minipage}{.55\linewidth}
\caption{Ablation for Effect of Reconstruction (Recon.) Loss. Reconstruction during pre-training plays an important role in aligning the MAE to extract meaningful information.}
\begin{tabu}{lccccc}
    \toprule
      & Stage 1 & Stage 2 & VQAv2 & GQA & OKVQA  \\
      & Recon. & Recon.  &  &  &  \\
    \midrule
    & &  & 33.1 & 25.4 & 12.1 \\
    & \cmark & \cmark & 52.4 & 32.2 & 29.2  \\
    Ours & \cmark  &  & \textbf{55.0}  & \textbf{34.9} & \textbf{34.2}  \\
    \bottomrule
  \end{tabu}
  \label{tab:abl_recon}
  \end{minipage}
\end{table}

\subsection{Ablation Analysis}
\label{sec:abl}
We perform the following ablations to analyze the various components of our approach. Please refer to Supplementary Section 4 for more ablation analysis.

\paragraph{\textbf{Leveraging Early Layer CLIP features}}
To illustrate the efficacy of MAE embeddings in learning better local representations, we conduct experiments using features from intermediate layers of CLIP. Specifically, we explore the utilization of features from layers $26, 28, 30$ as substitutes for MAE features  in our proposed approach. It is important to note that for this training, there is no reconstruction loss since we are employing CLIP features. 
From Table~\ref{tab:abl_interim}, it can be seen that employing MAE features leads to best results. We present more results in Supplementary Section 4.

\paragraph{\textbf{Impact of Image Reconstruction Loss in Pre-training.}} We examine the influence of the reconstruction loss during the pre-training stage.
In this pre-training phase, we employ four objectives: image-text contrastive and matching loss (for discriminative vision-language alignment), reconstruction loss, and text generation loss.
Our findings demonstrate that combining alignment and reconstruction objectives during pre-training, the image reconstruction loss becomes effective in extracting aligned and meaningful representations. To illustrate this, we conducted an experiment without the MAE decoder and reconstruction objective during pre-training%
and computed MAE features for the entire image without any masking. As shown in Table~\ref{tab:abl_recon}, row 1, the significant performance drop highlights the crucial role of the reconstruction loss in aligning and extracting useful information from MAE. %
This suggests that network cannot find a shortcut thus leading to drop in performance while MIM enables extracting meaningful and aligned representations leading to best performance.

\paragraph{\textbf{Impact of Image Reconstruction Loss in LLM Alignment}}
 We investigate the influence of reconstruction loss in stage-2. In the context of LLM alignment, the model undergoes training with language-modeling loss exclusively, without the inclusion of image-text contrastive and matching losses, and Q-Former does not receive any text-input. It is evident that combining only language modeling loss and reconstruction loss yields suboptimal results, as indicated by the performance drop shown in Table~\ref{tab:abl_recon}, row 2.

\section{Related Works}
\label{sec:rel_works}%
\paragraph{\textbf{Multimodal Large Language Models (MLLMs)}}
The success of Large Language Models (LLMs) has prompted researchers to delve into the exploration of integrating visual components into these models, culminating in the development of Multimodal LLMs (MLLMs) \cite{yin2023survey,li2023large}. 
MLLMs have garnered significant traction in both academic and industrial spheres due to their remarkable proficiency in comprehension and generation. 
The key idea is to leverage off-the-shelf pre-trained vision encoders and LLMs and keep them frozen during the training.
However, the most critical challenge in utilizing a frozen LLM lies in narrowing down the gap between visual features and the text space.
Existing MLLMs can be broadly divided into three categories according to the modules/components they used for bridging the modality gap: (i) Perceiver-based \cite{jaegle2021perceiver,jaegle2021perceiverio}, (ii) Q-Former-based \cite{li2023blip2}, and (iii) linear projection layer-based.
In Perceiver-based methods such as Flamingo \cite{alayrac2022flamingo}, they employ a Perceiver Resampler to produce a small fixed number of visual tokens per
image, subsequently amalgamating them with text tokens as input for LLMs. 
In other words, the Perceiver Resampler relies on the image-to-text generative learning to bridge the modality gap.
Q-Former shares the similar spirits with the Perceiver Resampler, except that Q-Former relies on an extra vision-language representation learning stage. 
Owing to its simplicity and efficiency, Q-Former is widely used in such as BLIP-2 \cite{li2023blip2}, SEED \cite{ge2023planting}, MiniGPT-4 \cite{zhu2023minigpt}, and InstructBLIP \cite{instructblip}. 
In linear projection layer-based methods, the common practice is to align visual features with text features through a singular linear layer before incorporating them into LLMs.
The effectiveness of this simple strategy is evidenced by recent studies such as LLaVa \cite{liu2023llava} and FROMAGe \cite{koh2023grounding}.
Our work is inspird by Q-Former but with the following differences: (i) we extend Q-Former to handle two off-the-shelf vision encoders, i.e., CLIP-ViT and MAE-ViT, and (ii) we introduce multimodality-to-multimodality generative learning to further bridge the modality gap.

\paragraph{\textbf{Self-Supervised Vision Encoders}}
Self-supervised vision encoders (VEs) play a crucial role in MLLMs by providing visual features that are understandable by LLMs.
Among them, VEs that are pre-trained by vision-language-based contrastive learning (CL) has been the most popular one, where the VE is trained to bring representations of matched image-text pair close together and push representations of unmatched pairs apart \cite{radford2021clip,yang2022vision,duan2022multi,chen2022we,jiang2023understanding,hu2023provla}. 
This encourages the VE to capture semantic similarities and differences in visual content.
However, recent work reveals that CL mainly focuses on low frequency signals and longer-range global patterns inheriting from its training objective \cite{park2023whatself}.
In consequence, CL-based MLLMs suffer from understanding detailed perceptions which are essential for tasks that require fine-grained visual understanding such as object counting.
As a counterpart, masked image modeling (MIM) involves masking parts of an image and tasking the vision encoder with predicting the masked image patches \cite{he2022mae}. 
This enhances the VE's ability to understand detailed %
visual features by promoting contextual understanding, encouraging the learning of spatial relationships, and facilitating the development of transferable representations.
Inspired by these observations, recent work attempt to build VEs that is able to understand both global semantic and detailed %
local patterns ~\cite{maenoscaleWeers_2023CVPR,park2023whatself}.
The key idea is to leverage the strength of CL and MIM by linearly combining two training objectives with a shared VE.
While simple and effective, these models are not readily applicable to MLLMs due to their sole pre-training on limited datasets and modest model sizes, significantly lagging behind their CL and MLM counterparts.
Although scaling up data and model size is possible, it introduces substantial carbon emissions and fails to capitalize on the advantages offered by off-the-shelf VEs from both CL and MIM.
In contrast, our approach incorporates a lightweight transformer that harnesses the benefits of pre-trained CL and MIM models, showcasing superior performance in fine-grained perception understanding without imposing a significant computational burden.

\section{Conclusion}
\label{sec:summary}
In this paper, we introduce X-Former, a novel architecture designed to enhance visual representations for Multimodal Language Models (MLLMs) by integrating pre-trained MAE and CLIP vision encoders. Our motivation stems from several observations: (i) existing MLLMs primarily rely on CLIP-ViT, which often fails to capture fine-grained visual signals; (ii) our empirical studies reveal that simply combining CLIP-ViT and MAE-ViT does not necessarily yield performance improvements; and (iii) the efficacy of MLLMs heavily depends on large-scale image-text pairs for pre-training and meticulously curated instruction tuning datasets for fine-tuning. X-Former effectively tackles these limitations by integrating CLIP-ViT and MAE-ViT through a dual cross-attention mechanism, all while keeping computational demands manageable. Our approach is plug-and-play and can be applied to other models. 
Our experimental results unequivocally show that X-Former surpasses BLIP-2 in a variety of visual reasoning tasks requiring robust visual comprehension. Remarkably, these superior results are achieved using only one-tenth of the image-text pair dataset, without the need for any instruction tuning datasets.

\clearpage

\noindent  \textbf{Supplementary Material for X-Former: Unifying Contrastive and Reconstruction Learning for MLLMs}

\vspace{2em}

\setcounter{section}{0}

\noindent The supplementary material is organized as follows. First, we provide additional details in Section~\ref{sec:dt} along with computational cost, zero-shot retrieval and parameter comparison in Section~\ref{sec:dt_comp},~\ref{sec:dt_zs_ret},~\ref{sec:dt_param} respectively. 

In Section~\ref{sec:ft}, we present the fine-tuning experiments followed by the Large-scale experimental results in Section~\ref{sec:largescale}. We report additional ablation analysis in Section~\ref{sec:abl_add}, and present additional qualitative results in Section~\ref{sec:morequal}. Finally, we present fine-grained query analysis in Section~\ref{sec:quer_div} and additional fine-grained evaluations in Section~\ref{sec:more_fg_res}. 

\section{Details}
\label{sec:dt}

\paragraph{\textbf{Implementation Details}} Our model undergoes pre-training for nine epochs in Stage-1 and one epoch in Stage-2, with OPT employed for Stage-2 alignment. Consistent with~\cite{li2023blip2}, we opt to utilize the output features from the second-to-last layer of CLIP-ViT and the parameters of the frozen ViTs and LLMs are converted into FP16. The AdamW optimizer with $[\beta_1 = 0.9$, $\beta_2 = 0.98]$ and a weight decay of $0.05$ is used. We use cosine learning rate decay with a peak learning rate of 1e-4 and a linear warmup of 2k steps. Images are resized to $224\times224$ with random resized crop and horizontal flip augmentations applied. The masking ratio for MAE-ViT is set to 50\%. During Stage-2 training, the minimum learning rate is maintained at 5e-5. 

Note that BLIP-2 does not release the captions utilized for training. Therefore we use the WebCapFilt captions from BLIP~\cite{li2022blip} for the LAION115M, SBU, and Conceptual Captions datasets, each of which contains one synthetic caption per image.
For BLIP-2 training, the authors perform further processing using CapFilt method to generate $10$ synthetic captions per image. These captions, along with the original caption, are ranked using CLIP ViT-L/14 image-text similarity, and the top $2$ captions per image are retained as training data. During training, one caption is randomly selected.

For fair comparison, we train BLIP-2 and our model on same training dataset and report results. Additionally, we present results with $FlanT5_{XL}$~\cite{chung2022scalingT5} model, which is trained with a prefix language modeling loss as mentioned in BLIP-2. For further details, please refer to Section~\ref{sec:largescale}. 

For a fair comparison with official BLIP-2 model, we take the official BLIP-2 checkpoint and use the provided evaluation script to report results as highlighted by * in \textit{all} the tables.

 \setlength{\columnsep}{7pt}%
 \setlength{\tabcolsep}{4pt}
\begin{table}
  \caption{\textbf{Computational Cost.}}
  \centering
  \begin{tabular}{lcccccc}
  \toprule
   \textbf{Method} & Train & \multicolumn{2}{c}{GPU Mem} & \multicolumn{2}{c}{\#{\sc flops}} & Inference \\
   \cmidrule(lr){3-4} \cmidrule(lr){5-6}
     & time & S1 & S2 & S1 & S2 & time \\
    \hline
    BLIP-2 & 39 hrs & 3.3G & 21G & 3.08T & 17.12T & $\sim$680 ms \\
    \textbf{X-Former (Ours)} & 43 hrs & 4.6G & 22G & 3.16T & 17.2T & $\sim$890 ms \\
    \hline
  \end{tabular}
  \label{tab:comp_cost}
\end{table}

\subsection{Computational Cost}
\label{sec:dt_comp}
Here, we discuss the training and GPU memory usage of our model compared to BLIP-2. Our method uses $4.7\%$ more GPU memory than BLIP-2 with $10\%$ higher train time for $OPT_{6.7B}$ model (Table~\ref{tab:comp_cost}). Note that vision encoders have much less params compared to LLMs, hence adding a vision encoder does not add much overhead. We also present detailed comparison between BLIP-2 and our model for both stage 1 and stage 2 in Table~\ref{tab:comp_cost}.

\begin{table}\setlength{\tabcolsep}{6pt}
\setlength{\columnsep}{7pt}%
  \centering
    \centering
    \caption{{\textbf{Zero Shot Retrieval Flickr}}}
    \begin{tabular}{c|cc|cc}
    \hline
    \textbf{Method} & TR@1 & TR@5 & IR@1 & IR@5 \\
    \hline    
    BLIP-2 & 89 & 98.3 & \textbf{83.5} & 96.2 \\
    \textbf{X-Former (Ours)} & \textbf{91.4} & \textbf{98.7} & 83.3 & \textbf{96.3} \\
    \hline
    \end{tabular}
    \label{tab:zs_ret}
\end{table}

\subsection{Zero-Shot Retrieval Results}
\label{sec:dt_zs_ret}
 image-text retrieval on Flickr dataset. Note that we use the pre-trained stage one model without any fine-tuning and compare with BLIP-2. As shown, our method improves retrieval scores over BLIP-2. 

\subsection{Detailed Parameter Comparison}
\label{sec:dt_param}
In Table~\ref{tab:abl_fuse_add}, we provide detailed comparison with other variants as discussed (see Section 2.2 in main text) to fuse MAE and CLIP features. Specifically, we provide number of trainable parameters along with performance on three VQA benchmarks namely VQAv2, GQA and OKVQA.
We show our approach achieves best performance with a gain of $2.8\%$ (Concat), $2.2\%$(Early CA) on GQA. In terms of parameters, Early CA adds $75$M more params than BLIP-2; $53$M more params than ours (Table~\ref{tab:abl_fuse_add} row 3). Despite having more params the performance is inferior to ours by margins of $1.2\%$ (VQA), $2.2\%$(GQA), $2.7\%$ (OKVQA).
\begin{table}[htb]
\caption{{\textbf{Detailed Comparison}. CA, C, M denote Cross-Attention, CLIP and MAE respectively}}
  \centering
  \resizebox{\textwidth}{!}{
  \begin{tabular}{lcccccc}
    \toprule
       \textbf{Method} & Model Details & Input & \# Trainable Params & VQAv2 & GQA & OKVQA  \\
    \midrule
    BLIP-2 & Q-Former & C & 108M & 52.4 & 33.1 & 31.5 \\
    Concat & Q-Former & M, C & 110M & 52.3 & 32.1 & 31.9  \\
    Early CA & Q-Former (M - CA) & M, C & 183M & 53.8 & 32.7 & 31.5 \\
    \textbf{X-Former (Ours)} & X-Former & M, C & 130M & \textbf{55.0} & \textbf{34.9} & \textbf{34.2}  \\
    \bottomrule
  \end{tabular}
  }
  \label{tab:abl_fuse_add}
\end{table}
\section{VQA Fine-tuning}
\label{sec:ft}
We have implemented the fine-tuning code for VQA task, as the authors have not released the code for this task. For Visual Question Answering fine-tuning task, we utilize VQA train and val splits along with Visual Genome train dataset following ~\cite{li2022blip, li2023blip2}. 

Note that the VQAv2 dataset contains multiple answer annotations per question. For our fine-tuning experiments, we randomly select one of the answers as the output for the VQA dataset.
As mentioned in BLIP-2~\cite{li2023blip2}, we feed the question as input to the X-Former along with the image embeddings for our model. %

It is important to highlight that BLIP-2 completely unfreezes the CLIP ViT, resulting in a total of $1.2$ Billion trainable parameters during fine-tuning. However, due to resource constraints, we are unable to train such large models. Therefore, we only unfreeze the layer norms in CLIP ViT while keeping the MAE ViT completely frozen, resulting in a total of 216 Million trainable parameters, which is $6 \times$ lower compared to full fine-tuning. We set the image size to 490 and the learning rate to 1e-5. During generation, we utilize the prompt "Question: ${}$ Short Answer:" and set the beam size to 5 for beam search.
Upon fine-tuning on the VQA task, our method achieves a 4.4\% improvement over the zero-shot performance for GQA dataset, which requires detailed visual understanding for accurate answers.

\section{Large Scale Results}
\label{sec:largescale}

\setlength{\tabcolsep}{4pt}
\begin{table*}[t]
    \centering
    \caption{\textbf{Zero-shot Visual Question Answering} results on GQA and OKVQA datasets with Large scale training. * evaluated using official checkpoint}
     
  \resizebox{0.96\textwidth}{!}{
  \begin{tabu}{lcccccc}
    \toprule
     & \multicolumn{2}{c}{Caption Pre-processing} &  &  & \\
       Method & CLIP Caption  & \#Caps/Img & \multirow{2}{1em}{Stage 1} & Stage2 & GQA  & OKVQA  \\
       & Ranking & & Data & Data & Acc. & Acc.\\
    \midrule
    Frozen~\cite{Frozen_NEURIPS2021_01b7575c} & & & & & - & 5.9 \\
    VLKD~\cite{VLKD_dai-etal-2022-enabling} & & &  & & - & 13.3 \\
    FewVLM~\cite{fewvlmjin-etal-2022-good} & & & & & 29.3 & 16.5 \\
    Flamingo3B~\cite{alayrac2022flamingo} & & &  & & - & 41.2 \\
    Flamingo9B~\cite{alayrac2022flamingo} & & &  & & - & 44.7 \\
    Flamingo80B~\cite{alayrac2022flamingo} & & &  & & - & 50.6 \\
    PNP-VQA T0$_{3B}$~\cite{tiong-etal-2022-plug} & & &  & & 32.3& 26.6 \\
    PNP-VQA T0$_{11B}$~\cite{tiong-etal-2022-plug} & & &  & & 33.4& 30.5 \\
    PNP-VQA UnifiedQAv2$_{3B}$~\cite{tiong-etal-2022-plug} & & &  & & 42.3 & 34.1 \\
    PNP-VQA UnifiedQAv2$_{11B}$~\cite{tiong-etal-2022-plug} & & &  & & 41.9 & 35.9 \\
    \rowfont{\color{gray}}
    BLIP-2 $OPT_{2.7B}$*~\cite{li2023blip2} & \cmark & 2 & 129M & 129M & 32.5 & 31.5  \\
    \rowfont{\color{gray}}
    BLIP-2 $FlanT5_{XL}$*~\cite{li2023blip2} & \cmark & 2 & 129M & 129M & 43.9 & 41.2  \\
    \hline
    BLIP-2 $OPT_{2.7B}$~\cite{li2023blip2} & \xmark & 1 & 14M & 105M & 32.2 & 25 \\
    \textbf{X-Former (Ours)} $\mathbf{OPT_{2.7B}}$ & \xmark & 1 & 14M & 105M & \textbf{34.3} & \textbf{27.6} \\
    
    BLIP-2 $FlanT5_{XL}$~\cite{li2023blip2} & \xmark & 1 & 14M & 105M & 42.9 & 38.2 \\
    \textbf{X-Former (Ours)} $\mathbf{FlanT5_{XL}}$ & \xmark & 1 & 14M & 105M & \textbf{44.9} & \textbf{39.5}\\
    \bottomrule
  \end{tabu}}
  \label{tab:zr_s2_large}
\end{table*}

Here, we present the results for large-scale training to illustrate that our model performance scales with data size. We utilize, LAION115M dataset in addition to the 14M dataset used in the paper. Note that, the downloaded dataset using web urls has an approximate $20\%$ miss rate, leading to overall dataset size of $105M$ for large-scale training.

We employ the synthetic captions released by BLIP~\cite{li2022blip} for LAION115M, SBU and Conceptual Caption datasets. We use the Stage-1 model pre-trained with 14M dataset as weight initialization. The LLM alignment stage is trained for an epoch on the large-scale dataset. We experiment with $OPT_{2.7B}$ LLM and $FlanT5_{XL}$~\cite{chung2022scalingT5} LLM and present Zero-shot visual question answering results on GQA and OKVQA datasets. $OPT$ is a decoder-only LLM while $FlanT5$ is enocder-decoder LLM. Following BLIP-2~\cite{li2023blip2}, for $OPT$ model we train with language modeling loss while $FlanT5$ model is trained with prefix language modeling loss i.e., caption is split into two parts: prefix and suffix. The prefix text along with visual representation forms input to LLM encoder and the suffix text is used as generation target for LLM decoder. A random value from start to middle of sentence is picked to divide the caption into two parts.

We demonstrate that our model outperforms BLIP-2~\cite{li2023blip2} at scale in Table~\ref{tab:zr_s2_large}. Specifically, our model achieves a 2.1\% gain on GQA dataset and 2.6\% gain on OKVQA dataset respectively with $OPT_{2.7B}$ LLM. We show similar gains using $FlanT5_{XL}$ LLM as well; our approach improves by 2\% on GQA dataset and 1.3\% on OKVQA dataset respectively. Note that PNP-VQA~\cite{tiong-etal-2022-plug} performance relies heavily on QA model specifically UnifiedQAv2 is a task-specific model pretrained for question answering, and OFA~\cite{wang2022ofa} trains visual encoders while we keep it frozen hence we do not compare with it.

\section{Ablation Analysis}
\label{sec:abl_add}
\begin{figure}
    \centering
    \includegraphics[width=0.6\textwidth]{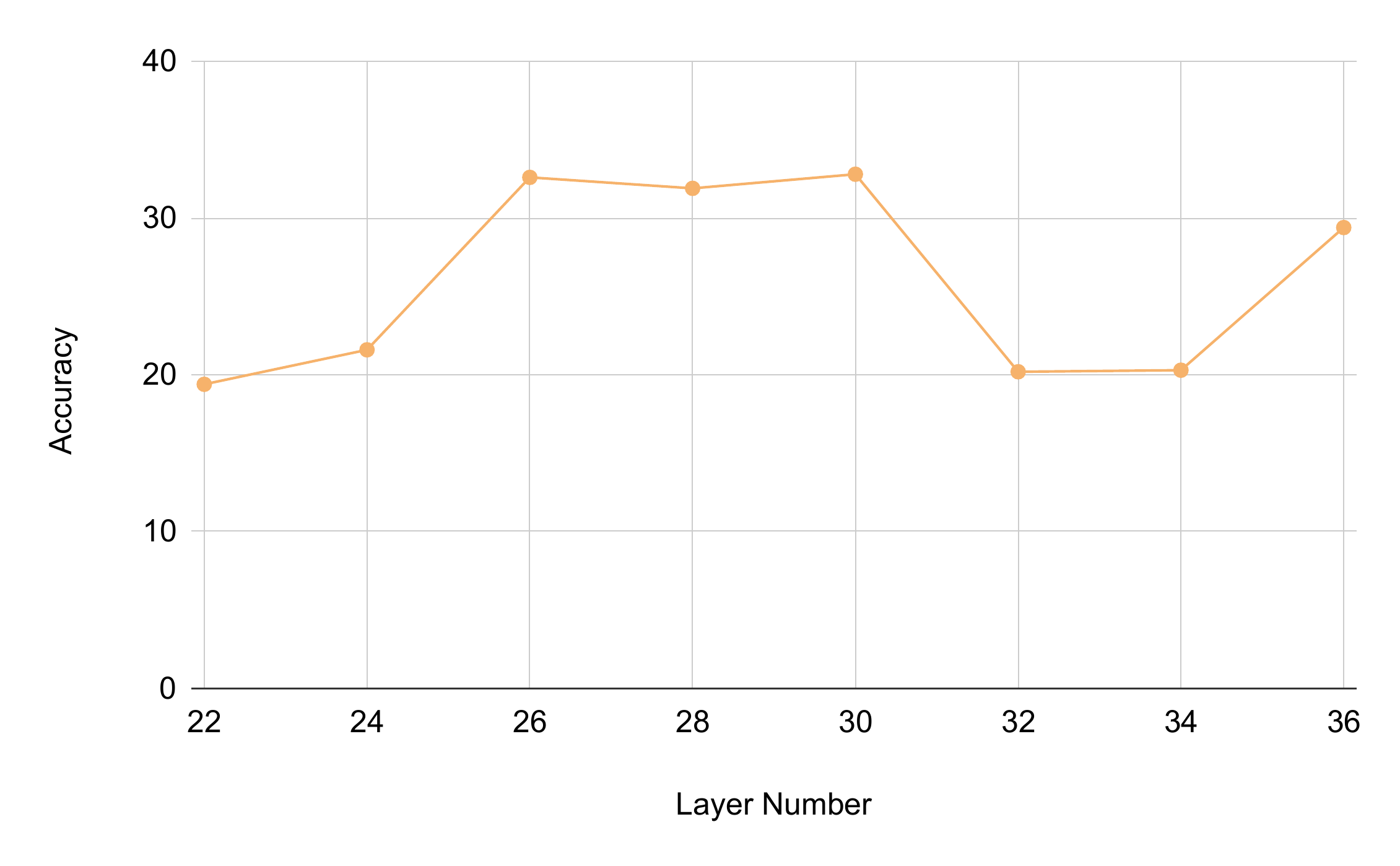}
    \caption{Zero-shot visual question answering performance on GQA datasets for different layer features from CLIP.}
    \label{fig:gqa_clip_layer}
  \end{figure}

\paragraph{\textbf{Ablation On CLIP Layers}}
As mentioned in Section 3.3 of main text ``Leveraging Early Layer CLIP features'',  we present additional results by experimenting with different layers from CLIP. We experiment with the following layers $\{22, 24,26,28,30,32,34,36\}$ as early layer features from CLIP ViT and report performance trend on GQA dataset. As shown in Figure~\ref{fig:gqa_clip_layer}, we observe that the best performance is achieved for layer 26 and layer 30, while utilizing features from layers below 26 leads to a drop in performance. Furthermore, using features from layers beyond 30 also results in a decline in performance.

Our findings demonstrate that the performance using early layer CLIP features is inferior to that of our model, with a 2\% decrease in performance compared to the best layer.

\paragraph{\textbf{More Ablations}}
We perform comprehensive ablations studies to analyze the impact of different loss components, effect of Horizontal flip augmentation and effect of Self-Attention. Further, we analyze the impact of X-Former training for LLM alignment and importance of MAE to capture detailed visual information complementing the global semantic representation from CLIP-ViT. Note that for these ablations, we use 8-A100swith batch size of 320/272 for stage 1 and stage 2 respectively.

 We find that ITG significantly affects retrieval more than ITM; without ITC, there is a slight drop in captioning performance. As shown in Table~\ref{tab:more_abl}, horizontal flip augmentation does not effect overall performance. For comprehensiveness, we analyze the effect of Self-Attention (SA) layer in X-Former as shown in Table~\ref{tab:more_abl}, row 5. There is a drop in captioning performance when we remove SA layer before the Cross-Attention with MAE.

\setlength{\columnsep}{4pt}%
 \setlength{\tabcolsep}{4pt}
\begin{table}[!htb]%
\centering
\caption{\textbf{Ablations Analysis.}}
    \label{tab:more_abl}
    \begin{tabular}{lcccccc}
    \hline
{\bf Method} & \textbf{TR5} & \textbf{TR10} & \textbf{IR5} & \textbf{IR10} & \textbf{B@4} & \textbf{C}\\
\hline
 w/o ITM & 96.6 & 98.8 & 93.8 & 96.7 & 35.9 & 120.3\\
 w/o ITG & 84.9 & 93.1 & 88.2 & 92.9 & - & - \\
 w/o ITC & - & -  & - & - &36.2 & 120.7 \\
 \hline
 w/o HFlip & 93.2 & 98.4 & 93.9 & 97.2 & 36.3 & 122.4 \\
 w/o SA & 95.5 & 99 & 93.9 & 97.1 & 35.6 & 120 \\
 \textbf{X-Former (Ours)} & 95.8 & 99 & 94 & 96.7 & 37 & 123.2\\
\hline
    \end{tabular}
\end{table}

\setlength{\columnsep}{4pt}%
 \setlength{\tabcolsep}{4pt}
\begin{table}[!htb]%
\centering
\caption{\textbf{Ablation Analysis.} *: smaller batch size in both stages}
    \label{tab:abl_s2}
    \begin{tabular}{lccccc}
    \toprule
       \textbf{Method} &  GQA & OKVQA  \\
    \midrule
    CLIP w Recon. & 22.5 & 8.1\\
    X-Former Frozen  & 25.5  & 15.9  \\
    \textbf{X-Former (Ours)} & \textbf{31.9} & \textbf{ 25.9} \\
    \bottomrule
  \end{tabular}
\end{table}

For LLM alignment, we follow BLIP-2 protocol and train X-Former in stage-2 along with a Fully Connected layer. To analyze the impact of training X-Former for LLM alignment, we experiment with frozen X-Former in stage-2 and report results in Table~\ref{tab:abl_s2} row 2. To demonstrate the importance of MAE further, we replace MAE encoder with CLIP-ViT and pass masked image to CLIP-ViT which is then optimized for image reconstruction with MAE decoder. As shown in Table~\ref{tab:abl_s2} row 1, the performance drops significantly by replacing MAE-ViT encoder with CLIP-ViT on both GQA and OKVQA dataset. Thus demonstrating MAE-ViT encoder plays crucial role in learning detailed visual features.

\section{Qualitative Results}
\label{sec:morequal}
\begin{figure*}
  \centering
    \begin{subfigure}[]{0.24\textwidth}
        \includegraphics[width=\textwidth]{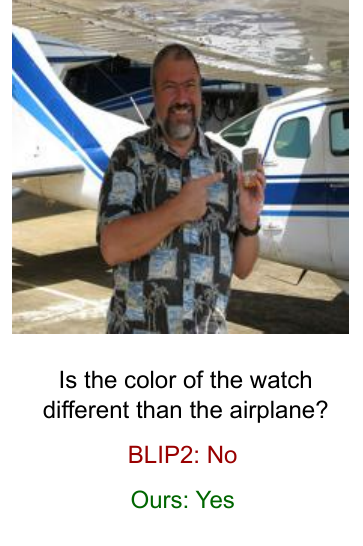}
    \caption{}
    \end{subfigure}%
    \begin{subfigure}[]{0.24\textwidth}
        \includegraphics[width=\textwidth]{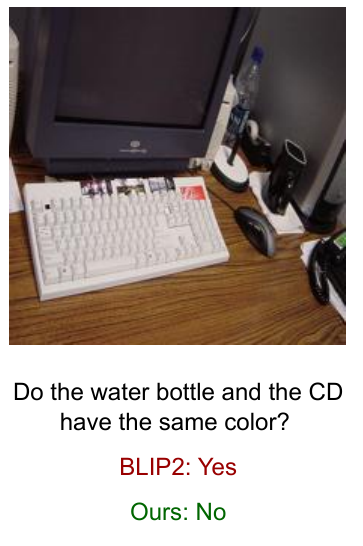}
        \caption{}
    \end{subfigure}
    \begin{subfigure}[]{0.24\textwidth}
        \includegraphics[width=\textwidth]{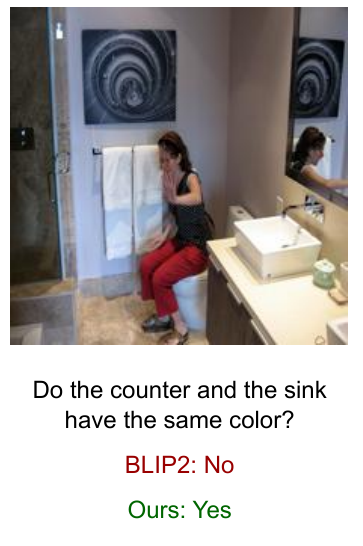}
        \caption{}
    \end{subfigure}
    \begin{subfigure}[]{0.24\textwidth}
        \includegraphics[width=\textwidth]{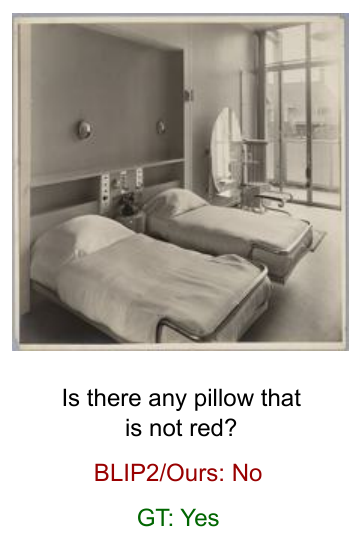}
        \caption{}
    \end{subfigure}
  \caption{Qualitative Comparison demonstrating ability to compare colors of specified objects.}
  \label{fig:qual_supp_color}
\end{figure*}

In this section, we present qualitative results, including cases where our method did not perform as expected. In Figure~\ref{fig:qual_supp_color}, we present examples that involve comparing the colors of different objects within the image. As you can see in Figure~\ref{fig:qual_supp_color} (a), (b), and (c), our method successfully understands the specified objects in the questions, regardless of their positions in the image, and compares their colors accurately. However, Figure~\ref{fig:qual_supp_color} (d) shows a more challenging scenario. Here, the pillow and the bed are not clearly distinguishable, which made it difficult for our model to identify the pillow in the image. 

\begin{figure*}
  \centering
    \begin{subfigure}[]{0.24\textwidth}
        \includegraphics[width=\textwidth]{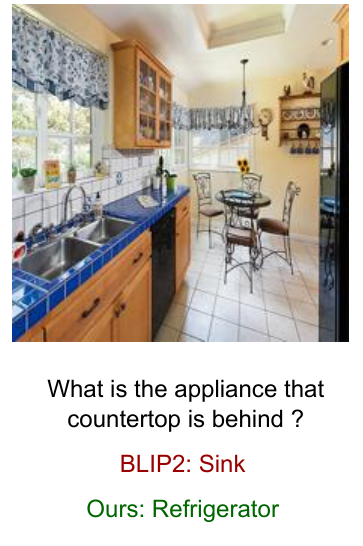}
    \caption{}
    \end{subfigure}%
    \begin{subfigure}[]{0.24\textwidth}
        \includegraphics[width=\textwidth]{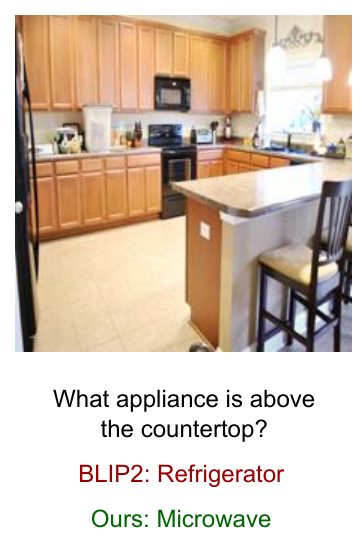}
        \caption{}
    \end{subfigure}
    \begin{subfigure}[]{0.24\textwidth}
        \includegraphics[width=\textwidth]{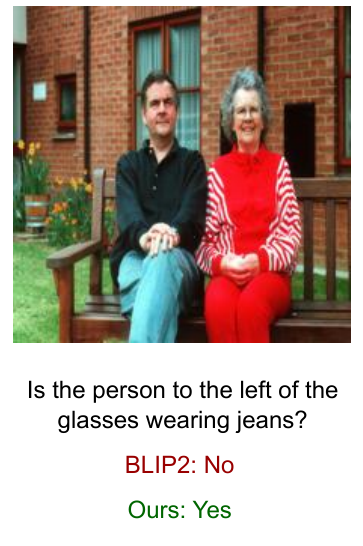}
        \caption{}
    \end{subfigure}
    \begin{subfigure}[]{0.24\textwidth}
        \includegraphics[width=\textwidth]{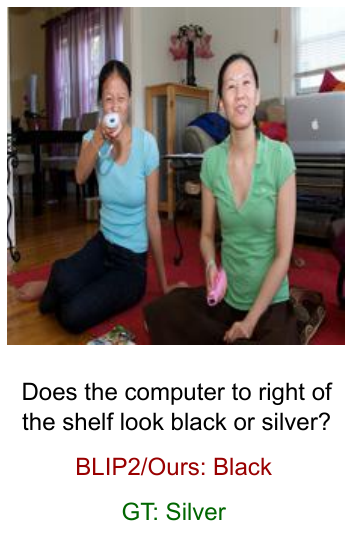}
        \caption{}
    \end{subfigure}
    
  \caption{Qualitative Comparison pertaining to question of spatial understanding.}
  \label{fig:qual_supp_sp}
\end{figure*}
Figure~\ref{fig:qual_supp_sp} showcases the spatial understanding capabilities of our model in comparison to the BLIP-2. A kitchen scene depicted in images ~\ref{fig:qual_supp_sp}(a) and (b), our model accurately identifies the refrigerator behind the countertop and the microwave above it, respectively. In contrast, BLIP-2 erroneously predicts a sink and refrigerator for the same questions. Our model correctly discerns the attire of individuals in ~\ref{fig:qual_supp_sp} (c), recognizing that the person to the left of the one wearing glasses is indeed wearing jeans—a detail that BLIP-2 overlooks. However, ~\ref{fig:qual_supp_sp} (d) presents a more challenging scenario for both models. When asked about the color of the computer to the right of the shelf, our model and BLIP-2 both incorrectly identify a silver laptop as black.

\begin{figure*}
  \centering
    \begin{subfigure}[]{0.24\textwidth}
        \includegraphics[width=\textwidth]{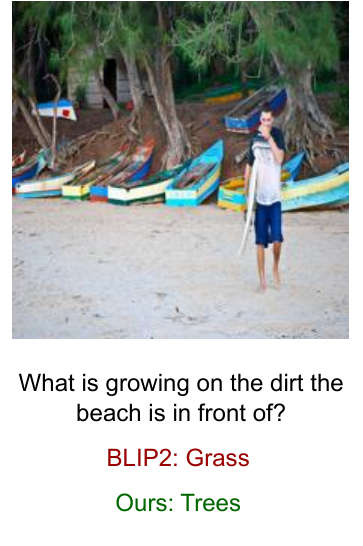}
    \caption{}
    \end{subfigure}%
    \begin{subfigure}[]{0.24\textwidth}
        \includegraphics[width=\textwidth]{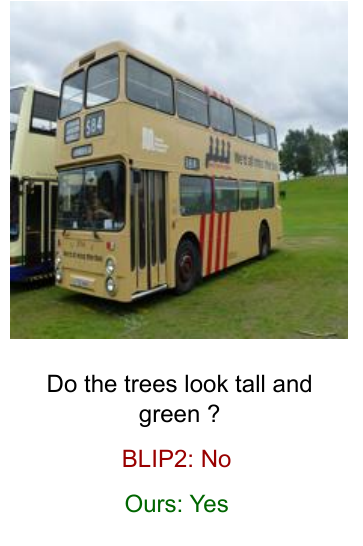}
        \caption{}
    \end{subfigure}
    \begin{subfigure}[]{0.24\textwidth}
        \includegraphics[width=\textwidth]{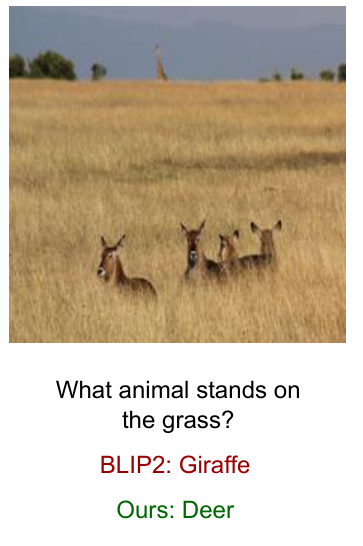}
        \caption{}
    \end{subfigure}
    \begin{subfigure}[]{0.24\textwidth}
        \includegraphics[width=\textwidth]{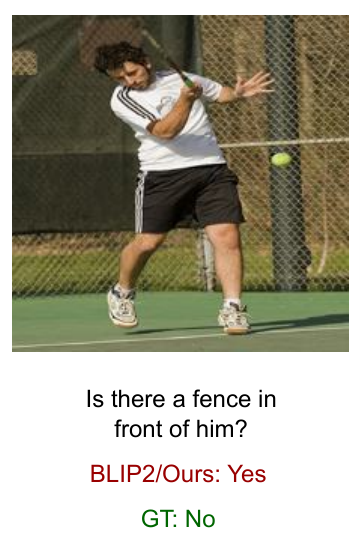}
        \caption{}
    \end{subfigure}
    
  \caption{Qualitative Comparison pertaining to question of relative object understanding in both background and foreground.}
  \label{fig:qual_supp_rel}
\end{figure*}

\begin{figure*}
  \centering
    \begin{subfigure}[]{0.24\textwidth}
        \includegraphics[width=\textwidth]{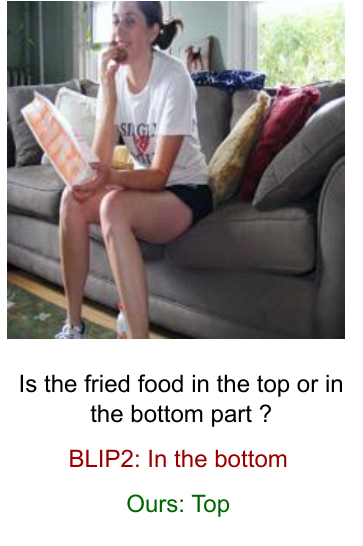}
    \caption{}
    \end{subfigure}%
    \begin{subfigure}[]{0.24\textwidth}
        \includegraphics[width=\textwidth]{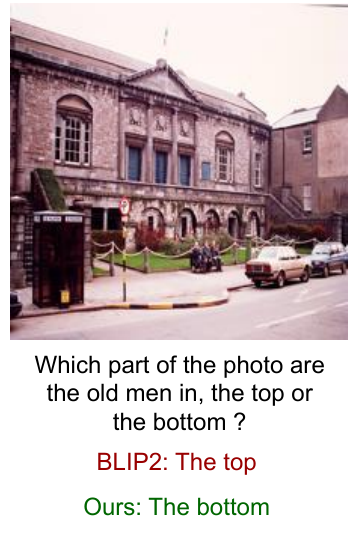}
        \caption{}
    \end{subfigure}
    \begin{subfigure}[]{0.24\textwidth}
        \includegraphics[width=\textwidth]{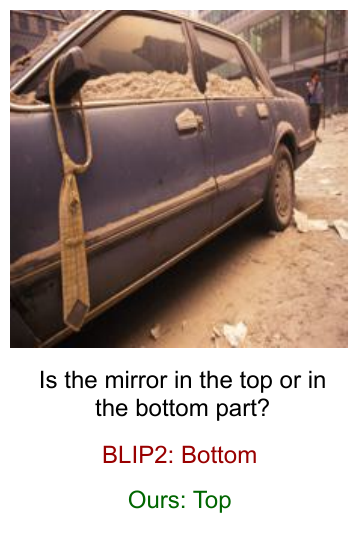}
        \caption{}
    \end{subfigure}
    \begin{subfigure}[]{0.24\textwidth}
        \includegraphics[width=\textwidth]{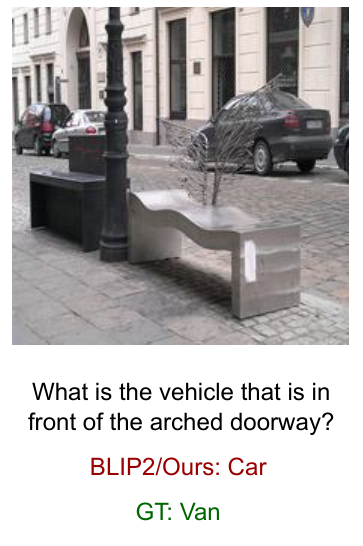}
        \caption{}
    \end{subfigure}
    
  \caption{Qualitative Comparison for questions relating to absolute image position understanding.}
  \label{fig:qual_supp_pos}
\end{figure*}

Figure~\ref{fig:qual_supp_rel} provides insights into the ability of our model to comprehend the relative positioning of objects within an image, both in the background and foreground. Figure~\ref{fig:qual_supp_rel} (a) and (b), probe the understanding of background elements, our model demonstrates a clear capacity to correctly identify objects, distinguishing trees on a beach and recognizing the tall, green trees beside a double-decker bus. This is in contrast to BLIP-2, which incorrectly identifies grass instead of trees and fails to acknowledge the verdancy and height of the trees. Further, in Figure~\ref{fig:qual_supp_rel} (c), which shifts the focus to foreground objects, our model accurately discerns the presence of deer in a grassy field. However, in Figure~\ref{fig:qual_supp_rel} (d), both our model and BLIP-2 inaccurately detect a fence in front of a tennis player, when, in fact, it is behind the player as shown. Overall, our model shows enhanced understanding of object contexts and positioning.

\begin{figure*}
  \centering
    \begin{subfigure}[]{0.24\textwidth}
        \includegraphics[width=\textwidth]{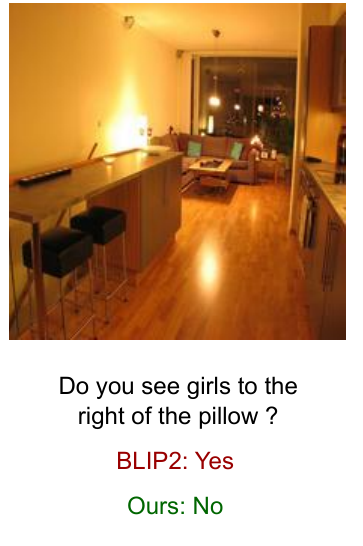}
    \caption{}
    \end{subfigure}%
    \begin{subfigure}[]{0.24\textwidth}
        \includegraphics[width=\textwidth]{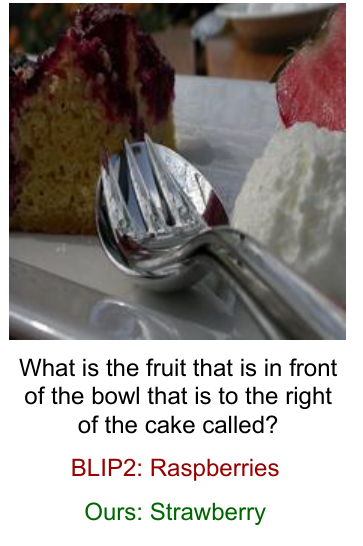}
        \caption{}
    \end{subfigure}
    \begin{subfigure}[]{0.24\textwidth}
        \includegraphics[width=\textwidth]{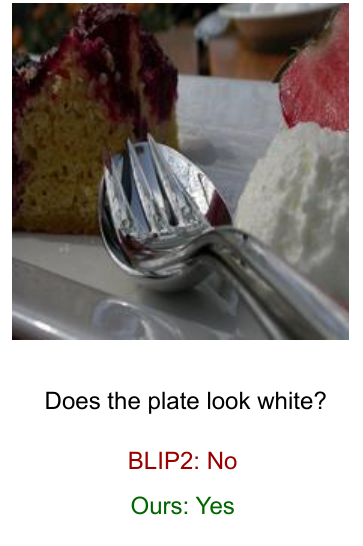}
        \caption{}
    \end{subfigure}
    \begin{subfigure}[]{0.24\textwidth}
        \includegraphics[width=\textwidth]{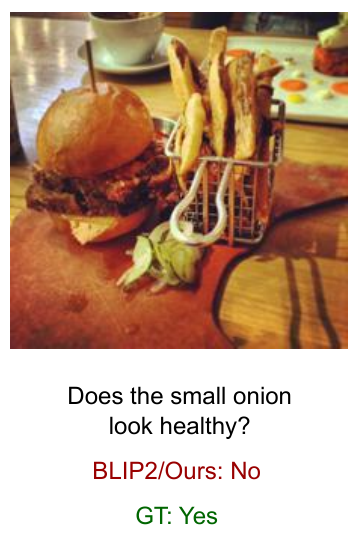}
        \caption{}
    \end{subfigure}
    
  \caption{Qualitative Comparison for samples with objects in close proximity in the scene.}
  \label{fig:qual_supp_cl}
\end{figure*}

Figure~\ref{fig:qual_supp_pos} exemplifies the absolute position reasoning capabilities of our model by assessing its ability to identify objects and their locations within an image, whether they are situated at the top or bottom parts of the image. In Figure~\ref{fig:qual_supp_pos} (a), (b) and (c), our model accurately determines the position of the fried food, the location of the old men, and the placement of the mirror, respectively, demonstrating better understanding of absolute positions within various contexts. However, Figure~\ref{fig:qual_supp_pos} (d) introduces a more complex situation involving multiple vehicles parked. Our model encounters difficulty here, incorrectly identifying a van as a car due to its close resemblance to car in this image.

Figure~\ref{fig:qual_supp_cl} demonstrates the capacity of our model to discern fine details of objects in close proximity within an image. In the living room scene depicted in Figure~\ref{fig:qual_supp_cl} (a), when questioned about the presence of a girl to the right of a pillow, our model accurately confirms the absence of a girl, whereas BLIP-2 incorrectly asserts a presence. The image of a food plate, shown in Figure~\ref{fig:qual_supp_cl}(b) and (c), further probe the model's ability to understand foreground and background distinctions. Our model correctly identifies the fruit next to the cake as a strawberry, where as BLIP-2 incorrectly categorizes it as raspberries.
Additionally, our model successfully distinguishes the white color of the plate amidst the various food items placed upon it, indicating a enhanced visual perception capabilities.
However Figure~\ref{fig:qual_supp_cl} (d), presents a complex scenario involving an assessment of an onion's quality, both our model and BLIP-2 fail to correctly evaluate its healthiness. This highlights the challenge in assessing condition/quality of food which is subject to interpretation.

\begin{figure}[thb]
    \centering
    \includegraphics[width=0.8\linewidth]{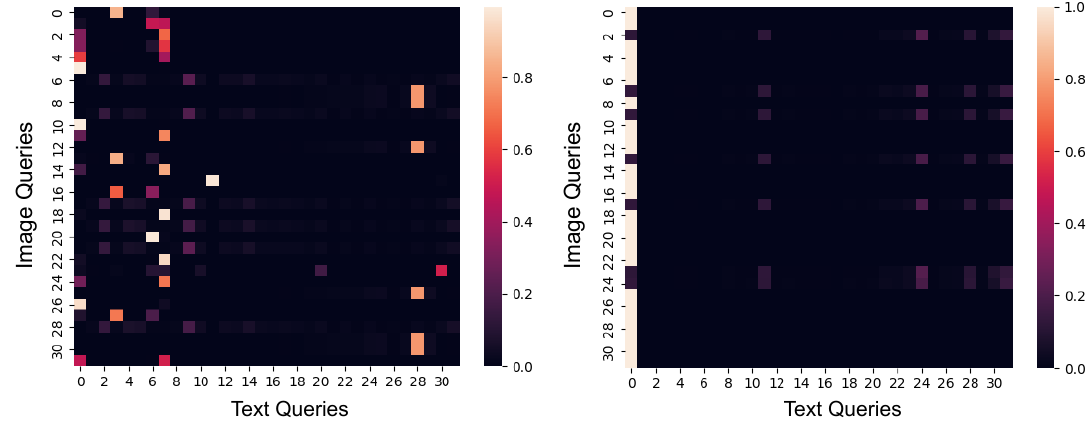}
    \caption{Query Diversity: Ours (left), BLIP-2 (right)}
    \label{fig:query_div}
\end{figure} 

\section{Query Diversity Comparison}
\label{sec:quer_div}
\begin{figure*}
  \centering
    \begin{subfigure}[]{0.8\textwidth}
        \includegraphics[width=\textwidth]{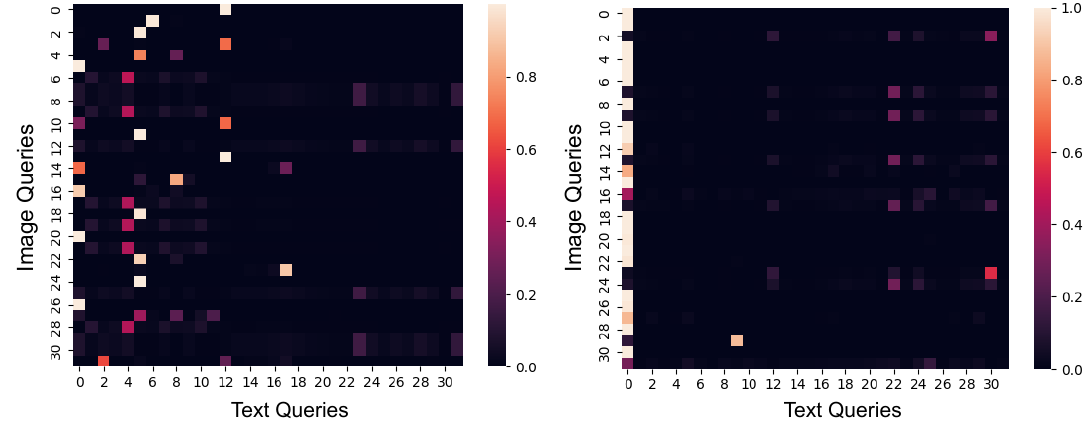}
    \caption{COCO\textunderscore val2014\textunderscore 000000001757.jpg}
    \end{subfigure}%
    \\
    \begin{subfigure}[]{0.8\textwidth}
        \includegraphics[width=\textwidth]{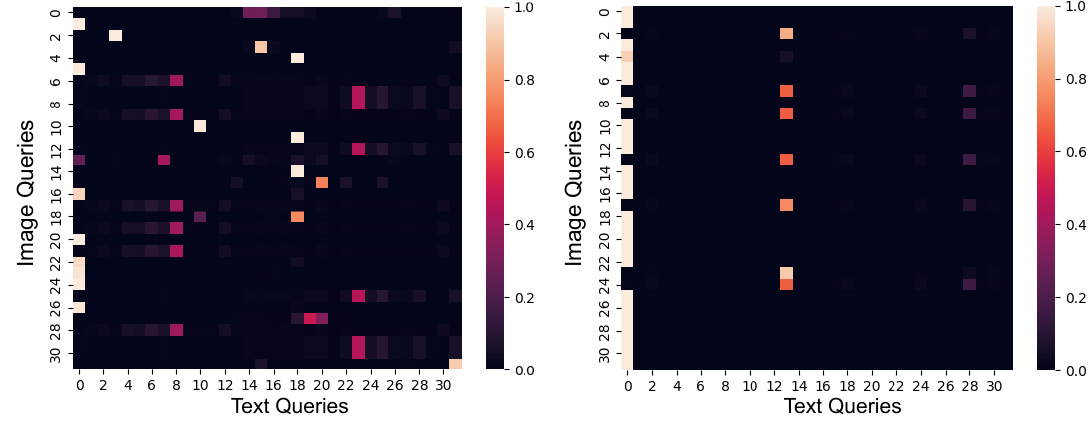}
        \caption{COCO\textunderscore val2014\textunderscore 000000031442.jpg}
    \end{subfigure}%
      \\
    \begin{subfigure}[]{0.8\textwidth}
        \includegraphics[width=\textwidth]{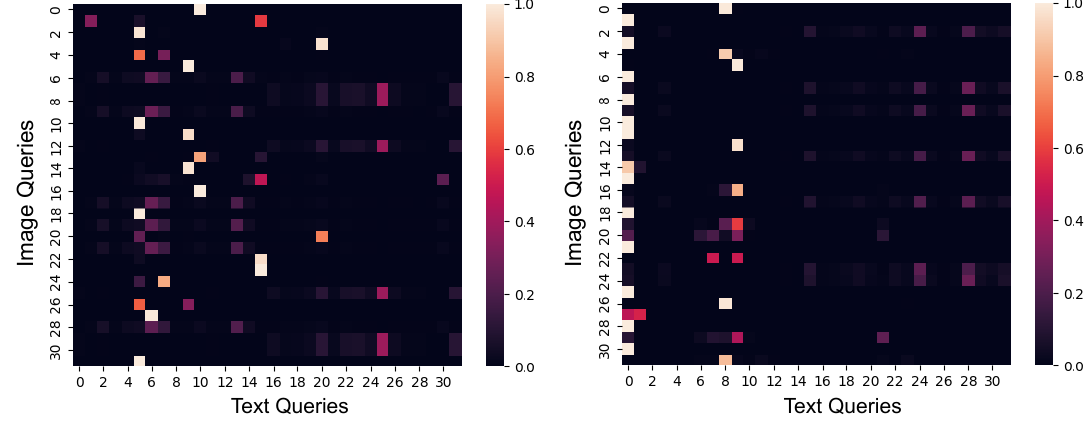}
        \caption{COCO\textunderscore val2014\textunderscore 000000002295.jpg}
    \end{subfigure}%
    \\
    \begin{subfigure}[]{0.8\textwidth}
        \includegraphics[width=\textwidth]{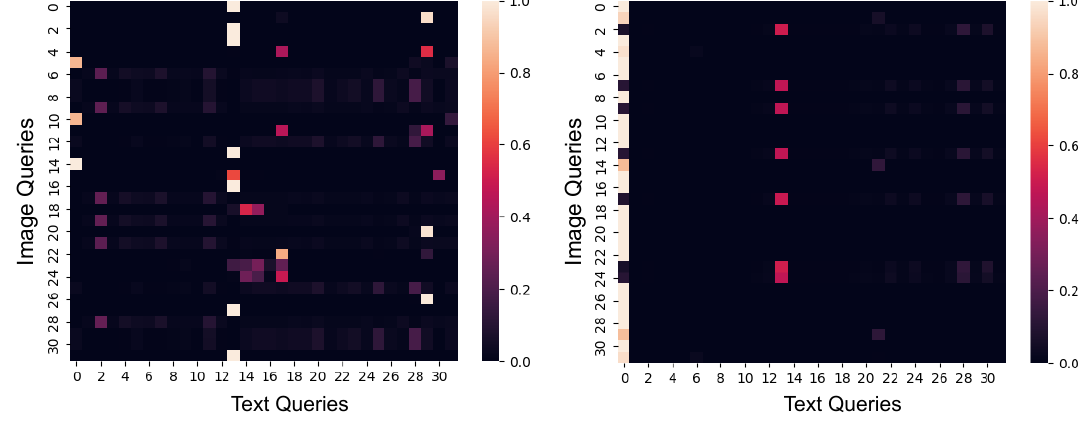}
        \caption{COCO\textunderscore val2014\textunderscore 000000002240.jpg}
    \end{subfigure}%
      \caption{Qualitative comparison of the queries for our model (left) with BLIP-2 (right). Our model learns diverse queries compared to BLIP-2. } 
  \label{fig:q_div_supp}
\end{figure*}

In addition to the fine-grained qualitative comparison, we further perform a fine-grained analysis of image-text queries from BLIP-2 and Ours. %
Particularly, we analyze the diversity of image-text query similarities as a proxy to investigate the fine-grained interaction between the image and text queries. For this, we first compute similarity between output queries for images and text using both Ours and BLIP-2 model to get $32\times 32$ matrix for each image-text pair as shown in Figure~\ref{fig:query_div}. We then aggregate the similarities by marginalizing them over the text to get scores for each image query.
To compute the overall diversity, we repeat this for all the samples in COCO karpathy test split and average it across samples and image queries. 
Our findings indicate that the queries learnt by our model are $7\%$ more diverse than those of BLIP-2, demonstrating the enhanced capability of our approach to capture a broader range of nuances in image-query representations. %
We present more fine-grained qualitative examples to show query diversity of our model compared to BLIP-2. As shown in Figure~\ref{fig:q_div_supp}, our model learns diverse queries than BLIP-2 for few samples from COCO karpathy test split.

\begin{table}[thb]
  \centering
  \caption{\textbf{Performance on SugarCrepe.}}
  \label{tab:screpe} 
  \begin{tabu}{lccccccc}
    \toprule
       \textbf{Method} & \multicolumn{3}{c}{\textbf{Object}} & \multicolumn{3}{c}{\textbf{Attribute}} & \textbf{Relation} \\
       \cmidrule(lr){2-4} \cmidrule(lr){5-7} 
        & Replace & Swap & Add &  Replace & Swap & Add & Replace  \\
    \midrule
    BLIP-2 & 93.4 & 56.7 & 89.1 & 81.9 & 66.9 & 83 & 72.1 \\
    \textbf{X-Former (Ours)} & \textbf{95.7} &\textbf{ 64.1} & \textbf{92.1} & \textbf{84.2} & \textbf{68.6} & 83 & \textbf{75.8} \\
    \bottomrule
  \end{tabu}
\end{table}

\setlength{\columnsep}{6pt}%
\setlength{\tabcolsep}{6pt}
\begin{table}
\centering
\caption{\textbf{Results on Flowers-102/Food-101}}
    \label{tab:fg_tasks}
    \begin{tabular}{lcc}
    \hline
 & \textbf{Flower102} & \textbf{Food101} \\
\hline
 BLIP-2 & 53.3 & 82.7\\
\textbf{X-Former (Ours)} &  \textbf{58.2} & \textbf{83.7}\\
\hline
    \end{tabular}
    \label{table:fg}
\end{table}

\section{Additional Fine-Grained Results}
\label{sec:more_fg_res}
OC/MCI~\cite{wang2023gvt} have been proposed in 2023 as benchmarks for measuring the
capability of MLLM in comprehending and reasoning about fine-grained visual features. Note that our usage of `fine-grained' refers to high-frequency and detailed visual representations that are overlooked by current MLLMs with CLIP-ViT as the visual backbone. It does not refer to traditional fine-grained vision tasks (e.g., bird species classification) that require fine-grained \textit{annotations}. 
We further evaluate our model for more fine-grained tasks, although our model is not dedicated to traditional ``fine-grained tasks". We evaluate on \textit{SugarCrepe (SC)}~\cite{hsieh2023sugarcrepe}, \textit{Flowers102}~\cite{nilsback2008flowers}, and \textit{Food-101}~\cite{bossard14food}. 
As shown in Tab.~\ref{tab:screpe},~\ref{tab:fg_tasks}, XFormer outperforms BLIP-2 by a large margin, indicating XFormer is also good at distinguishing visually similar objects.

\clearpage

\bibliographystyle{splncs04}
\bibliography{main}
\end{document}